\documentclass{article}

\usepackage[final]{tackling_climate_workshop_style}

\setcitestyle{numbers,square}
\usepackage[utf8]{inputenc}
\usepackage[T1]{fontenc}
\usepackage{hyperref}
\usepackage{url}
\usepackage{booktabs}
\usepackage{amsmath,amssymb,amsfonts}
\usepackage{nicefrac}
\usepackage{microtype}
\usepackage{graphicx}
\usepackage{subcaption}
\usepackage{multirow}
\usepackage{float} 

\title{Locked In: Multi-Objective RL for the \\
AI Data Center Build-Out}
\author{%
  \begin{tabular}[t]{c}
    \textbf{Satyapragnya Kar} \\
    \small New York University \\
    \small \texttt{sk13001@nyu.edu}
  \end{tabular}
  \hspace{0.5cm}
  \begin{tabular}[t]{c}
    \textbf{Rajarshi Das} \\
    \small AlphaInfra.ai, MQube Cognition \\
    \small \texttt{rajarshi@alphainfra.ai}
  \end{tabular}
  \hspace{0.5cm}
  \begin{tabular}[t]{c}
    \textbf{PR Panigrahi} \\
    \small Envana Software Solutions \\
    \small \texttt{pr.panigrahi@envana.com}
  \end{tabular}
}

\begin{document}
\maketitle

\begin{abstract}
AI data center build-out is a multi-trillion-dollar, multi-decade
infrastructure program. Yet siting, power procurement, and cooling decisions
are still made with static, single-point scoring tools. These tools are
evaluated once, on day-zero numbers. This is a poor fit. A site's water,
power, and carbon profile drifts substantially over the 10 to 20 year asset
life. The governing decisions are also irreversible, and they bind capital and
physical plant for years. We formalize long-horizon build-out as a tiered
multi-objective Markov decision process. Strategic commitments are gated to
discrete windows, incur switching costs, and energize only after
source-dependent lead times. Operational dispatch is restricted to capacity
that physically exists. We solve the problem with a preference-conditioned
multi-objective agent that has a semi-Markov manager and worker hierarchy.
The agent recovers a family of policies that a planner can steer across
cost, carbon, water, and reliability at inference time. We validate
long-horizon dynamics with a lightweight counterfactual rollout model.
Across three commitment regimes (\textsc{irreversible}, \textsc{reversible},
and \textsc{commit-once}) on real site data, preference conditioning yields a
confirmed, steerable trade-off on both objective pairs in every regime.
Imposing lock-in produces a more evenly covered frontier and a far more
consistent rate of anticipatory retrofit ahead of a binding water-stress
constraint than a freely reversible formulation. Finally, a compact
counterfactual dynamics model tracks slow-moving state well. It is measurably
outperformed by simple persistence on the water-stress axis. We report this
limitation rather than obscure it. We release the environment, agents,
baselines, and evaluation harness. All of them are grounded in real
interconnection-queue, water-stress, and grid-emissions data.
\end{abstract}

\vspace{-2pt}
\section{Introduction}
\vspace{-1pt}

Site selection and build-out planning for AI data centers is conventionally
treated as a one-shot scoring exercise. A candidate location is evaluated
against present-day electricity price, grid carbon intensity, and water
availability. It is then scored and filed. That approach is reasonable for a
slowly varying resource base. It is not reasonable here. Global data center
capital expenditure is projected to reach nearly \$7\,trillion by 2030
\citep{mckinsey2025}. Annual data center electricity demand is on track to
reach roughly 945\,TWh by 2030 \citep{iea2025energyai,goldman2025power}.
Roughly 2{,}060\,GW of generation and storage capacity sits in U.S.
interconnection queues, with typical multi-year waits to commercial operation
\citep{lbnl2025queued}. Water consumption is projected to grow by as much as
170\% over the same period \citep{westwater2025}. Hyperscaler capital
commitments already run into the hundreds of billions of dollars through
2027 \citep{cbre2025capex}. All-in build costs run from roughly
\$11\,million per MW up to the tens of millions once compute fit-out is
included \citep{jll2026outlook}. Together, these figures imply three things.
The resource base drifts within a single asset life. \emph{When} capacity
energizes is as consequential as \emph{what} is procured. A single build
phase is a multi-billion-dollar, effectively irreversible commitment. None of
this can be expressed by a score computed once.

Planning where, how, and when to build AI data center capacity (siting, power
procurement, and cooling) is a sequential decision problem, not a scoring
problem. There are two reasons. First, the objectives are contrastive. There
is no single optimum, only a Pareto frontier across cost, carbon, water, and
reliability. Second, decisions are irreversible on operationally meaningful
timescales, so early commitments constrain the entire remaining sequence. A
formulation that permits free, instantaneous switching collapses to myopic
per-period scoring. This is exactly the practice this work aims to displace.

We make three contributions. First, we introduce a tiered multi-objective MDP
for build-out planning. It has gated retrofit windows, switching costs, and
source-dependent energization lead times calibrated to real
interconnection-queue data (Section~\ref{sec:formulation}). Second, we
present a preference-conditioned agent with a semi-Markov manager and worker
hierarchy. The hierarchy separates decade-scale commitment from quarterly
dispatch while preserving inference-time steerability. We also provide a
lightweight counterfactual dynamics model for validating long-horizon
behavior (Section~\ref{sec:method}). Third, we report an empirical study
across three commitment regimes on real, grounded site data. The study
isolates the contribution of irreversibility to frontier quality,
steerability, and anticipatory behavior (Section~\ref{sec:results}). Related
work on RL for computing and energy systems, multi-objective RL, hierarchical
RL, and robust or non-stationary RL is summarized in
Appendix~\ref{app:related}.

This paper is one of a small family of companion studies on the AI build-out.
In \emph{Methane First} \citep{das2026methane}, we study a narrower,
single-decision slice of the same planning problem. The question is how a
fixed annual budget should be split between methane leak repair and carbon
capture across a fleet of on-site gas-powered data centers. That problem is
small enough to solve exactly by dynamic programming. A model-free RL policy
recovers the exact optimum in nearly all runs. This supports using RL on the
richer, multi-objective, multi-commitment setting studied here, where no exact
solution is available (Appendix~\ref{app:related}). A second companion,
\emph{Watts If?} \citep{chouhan2026wattsifbenchmarkcounterfactual}, benchmarks language models on
counterfactual reasoning about what each commitment in a build-out rules out
(see Section~\ref{sec:discussion}).

\vspace{-2pt}
\section{Problem Formulation}
\label{sec:formulation}
\vspace{-1pt}

\paragraph{Setup:} We model build-out as a multi-objective MDP
$\langle \mathcal{S}, \mathcal{A}, T, \mathbf{R}, \gamma \rangle$. The reward
is vector-valued, $\mathbf{R}: \mathcal{S}\times\mathcal{A} \to
\mathbb{R}^4$, with $\mathbf{R} = [-\text{cost}, -\text{carbon}, -\text{water},
+\text{reliability}]$. It is never scalarized inside the environment, so a
Pareto frontier remains recoverable. The base timestep is one quarter. The
horizon is 80 quarters (20 years). Full state, action, and site detail is in
Appendix~\ref{app:state}.

\paragraph{Decision tiers and irreversibility:} A \emph{manager} makes two
kinds of decisions. It selects the cooling architecture (air, liquid, or
immersion), which is legal only at discrete retrofit windows. It also selects
an annual procurement commitment among four power pathways. These are
behind-the-meter gas, clean firm power, on-site renewables, and grid supply.
They differ in cost, carbon, lead time, and reliability. A committed source
energizes only after a source-dependent lead time. This lead time is drawn
from calibrated interconnection-queue statistics. The switching cost is
charged against the cost objective. A \emph{worker} selects a quarterly
dispatch mix. The mix is masked so that it draws only on capacity that has
already energized.

We study three commitment regimes. They vary only in how often the manager's
cooling window opens. In the \textsc{irreversible} regime, a window opens
every 12 quarters, at a facility already operating evaporative cooling. In the
\textsc{reversible} regime, a window opens every quarter. This recovers the
free-switching setting implicit in static scoring. In the \textsc{commit-once}
regime, a window opens only at $t=0$. This is pure site and architecture
selection, with no later retrofit at all.

\paragraph{Objective:} We learn a family of policies $\{\pi_{\mathbf{w}}\}$
indexed by preference vectors $\mathbf{w}$ on the simplex. The family traces
the frontier of expected discounted vector returns. A planner can therefore
select an operating point at inference without retraining.

\vspace{-2pt}
\section{Method}
\label{sec:method}
\vspace{-1pt}

\paragraph{Preference-conditioned multi-objective agent:} A network
$Q_\theta(s,\mathbf{w}) \in \mathbb{R}^{|\mathcal{A}|\times 4}$ outputs
vector-valued action values. These are conditioned on the state and a sampled
preference vector. The greedy action maximizes $\mathbf{w}^\top
Q_\theta(s,\mathbf{w})_a$ over legal actions. Illegal actions are masked at
both selection and bootstrap. Preferences are resampled every episode, so one
network covers the frontier.

\paragraph{Semi-Markov hierarchy:} The manager acts every $K=4$ quarters. It
bootstraps over its commitment interval,
$y = \mathbf{R}^{(K)} + \gamma^{K} Q_{\bar\theta}(s',\mathbf{w})_{a^*}$. The
worker acts each quarter. It is conditioned on the state, the preference, and
the manager's active commitment, and it bootstraps at $\gamma$. Both levels
are preference-conditioned, so steerability is preserved through the
hierarchy. The coupling is structural, because manager commitments determine
which dispatch mixes are feasible for the worker. A \emph{flat} ablation
trains an otherwise identical agent over the joint $12\times7{=}84$-action
space. It differs only in lacking timescale separation and semi-Markov
bootstrapping.

\paragraph{Learned counterfactual model (LCM):} A learned counterfactual model
(LCM) is a learned simulator of the build-out environment. It predicts how a
site's state and objectives would evolve under a different sequence of
commitments. A 20-year policy cannot be validated against realized outcomes
within a research cycle. We therefore train a compact deep ensemble as our
LCM. It predicts next-state deltas and reward vectors under counterfactual
joint actions. This lets us ask ``what would have happened under a different
action sequence'' cheaply, as a diagnostic. The LCM is not used inside
training. We validate it with rolling-origin backtesting against persistence
and linear-drift baselines. We treat it as unsuitable for planning on any
dimension where it does not clearly beat both baselines. The full
architecture and validation protocol are in Appendix~\ref{app:lcm}.

\vspace{-2pt}
\section{Experimental Setup}
\vspace{-1pt}
We use six real site profiles (Appendix~\ref{app:state},
Table~\ref{tab:sites}). They span distinct water-stress, carbon-intensity,
price, and queue-pressure regimes. Each site is crossed with four
demand and climate scenarios. Four sites and three scenarios form the training
distribution. Two unseen sites and one unseen scenario are held out. All
methods are evaluated on an identical fixed sweep of 40 preference vectors
under common random numbers. They share a single hypervolume reference point.
We report hypervolume for the cost$\times$carbon and water$\times$reliability
views. We also report Pareto-front sparsity, where lower means more even
coverage. Preference-steerability correlations come from an eleven-point
sweep. Where a post-day-zero retrofit window exists, we report the rate at
which the policy leaves evaporative cooling ahead of a per-site water-stress
constraint. We call this the anticipation rate. Full data provenance is in
Appendix~\ref{app:data}.

\vspace{-2pt}
\section{Results}
\label{sec:results}
\vspace{-1pt}

\begin{table}[H]
\centering\small
\caption{Hypervolume (in-distribution / held-out) and sparsity, mean $\pm$
std over 3 seeds. cc is cost $\times$ carbon and wr is water $\times$
reliability. Anticipation rate is defined only where a post-day-zero retrofit
window exists. It is trivially zero under \textsc{commit-once}, so it is
omitted from this table (see Section~\ref{sec:results}).}
\label{tab:main}
\resizebox{\textwidth}{!}{%
\begin{tabular}{@{}llccc@{}}
\toprule
Regime & Agent & HV cc (in-dist / held-out) & HV wr (in-dist / held-out) & Sparsity \\
\midrule
\multirow{2}{*}{\textsc{irreversible}}
 & Flat         & $212.1\pm12.0$ / $14.0\pm10.3$  & $1142.7\pm6.7$ / $930.6\pm24.3$   & $9.38\pm0.84$ \\
 & Hierarchical & $222.6\pm6.7$ / $10.4\pm8.6$    & $1197.5\pm22.3$ / $1088.1\pm47.2$ & $9.58\pm0.75$ \\
\multirow{2}{*}{\textsc{reversible}}
 & Flat         & $209.0\pm19.2$ / $30.7\pm21.6$  & $1322.5\pm19.9$ / $1095.8\pm35.1$ & $25.43\pm15.44$ \\
 & Hierarchical & $250.8\pm8.8$  / $33.4\pm5.5$   & $1306.1\pm16.9$ / $1091.2\pm4.9$  & $16.80\pm2.02$ \\
\multirow{2}{*}{\textsc{commit-once}}
 & Flat         & $225.7\pm12.9$ / $47.3\pm15.9$  & $1312.8\pm18.7$ / $1023.3\pm100.6$ & $20.59\pm5.07$ \\
 & Hierarchical & $248.4\pm5.8$  / $46.3\pm5.5$   & $1294.7\pm4.0$  / $1098.9\pm34.9$  & $34.68\pm21.70$ \\
\bottomrule
\end{tabular}}
\end{table}

\paragraph{Hierarchy helps in most regimes, with one instructive exception:}
In both \textsc{irreversible} and \textsc{reversible}, the hierarchical agent
dominates its flat counterpart on both hypervolume views. For example, under
\textsc{irreversible} the cost$\times$carbon hypervolume is $222.6$ against
$212.1$. Under \textsc{commit-once} the picture is mixed. Hierarchy still
raises mean cost$\times$carbon hypervolume ($248.4$ vs.\ $225.7$). It also
lowers variance on three of the four hypervolume columns. However, it produces
markedly worse and less consistent sparsity than the flat agent
($34.68\pm21.70$ vs.\ $20.59\pm5.07$). This is the one cell in
Table~\ref{tab:main} where hierarchy does not clearly help. The result is
consistent with the regime's structure. There is only a single, unconstrained
day-zero choice, and no ongoing switching friction to coordinate. The
timescale separation that helps a hierarchical agent under lock-in or free
reversibility therefore has less to act on. Its benefit shows up mainly in
hypervolume stability rather than frontier evenness. Figure~\ref{fig:fronts}
shows the recovered fronts for the hierarchical agent under lock-in, free
reversibility, and \textsc{commit-once}. Flat-agent fronts and all learning
curves are in Appendix~\ref{app:figures}.

\newcommand{\frontw}{0.8\textwidth} 
\clearpage
\begingroup\captionsetup{type=figure}
{\centering\subcaptionbox{\textsc{irreversible}, hierarchical agent.}{\includegraphics[width=\frontw]{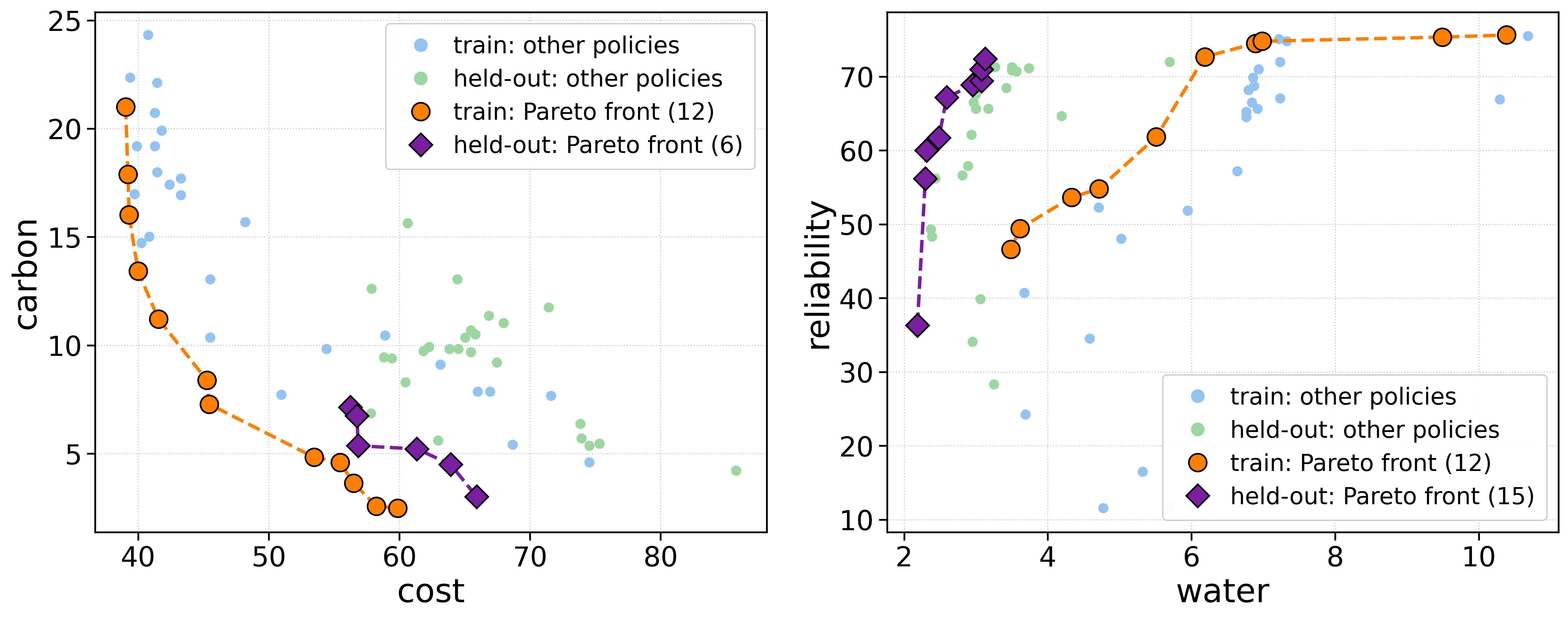}}\par}
\vspace{0.5em}
{\centering\subcaptionbox{\textsc{reversible}, hierarchical agent.}{\includegraphics[width=\frontw]{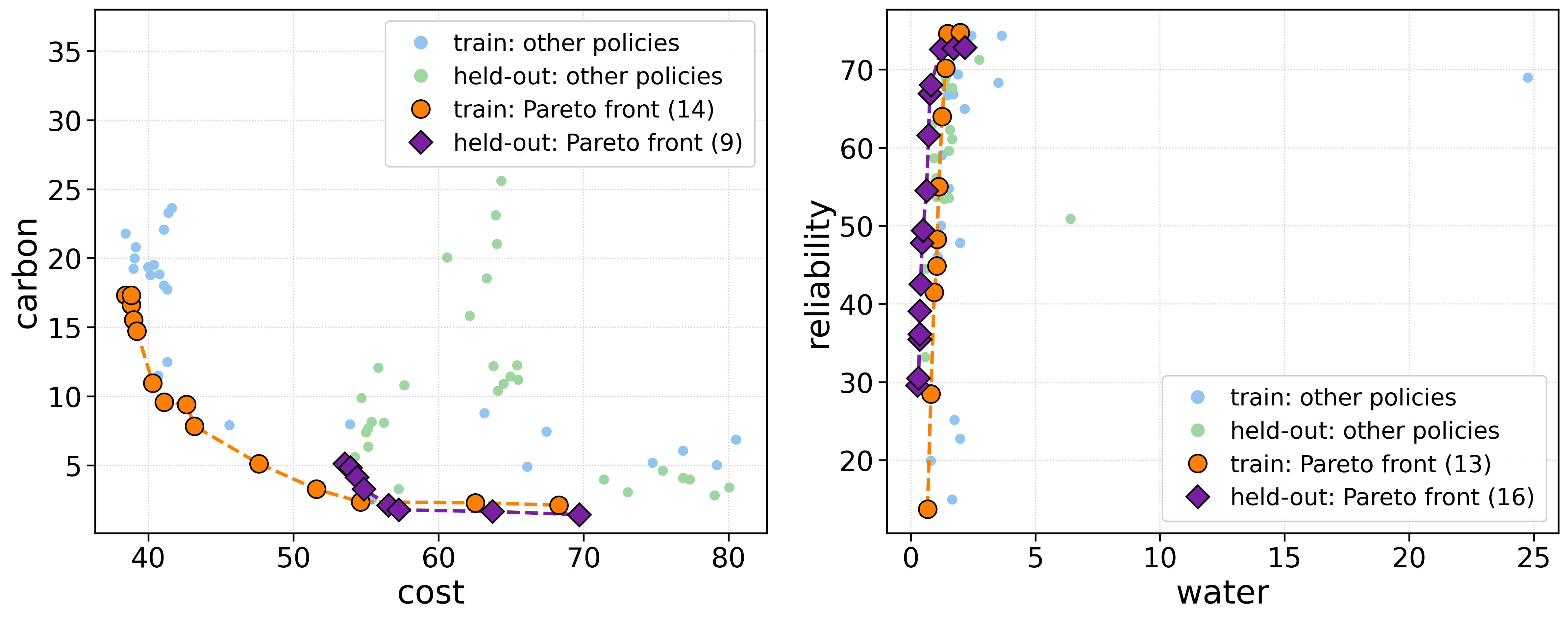}}\par}
\vspace{0.5em}
{\centering\subcaptionbox{\textsc{commit-once}, hierarchical agent.}{\includegraphics[width=\frontw]{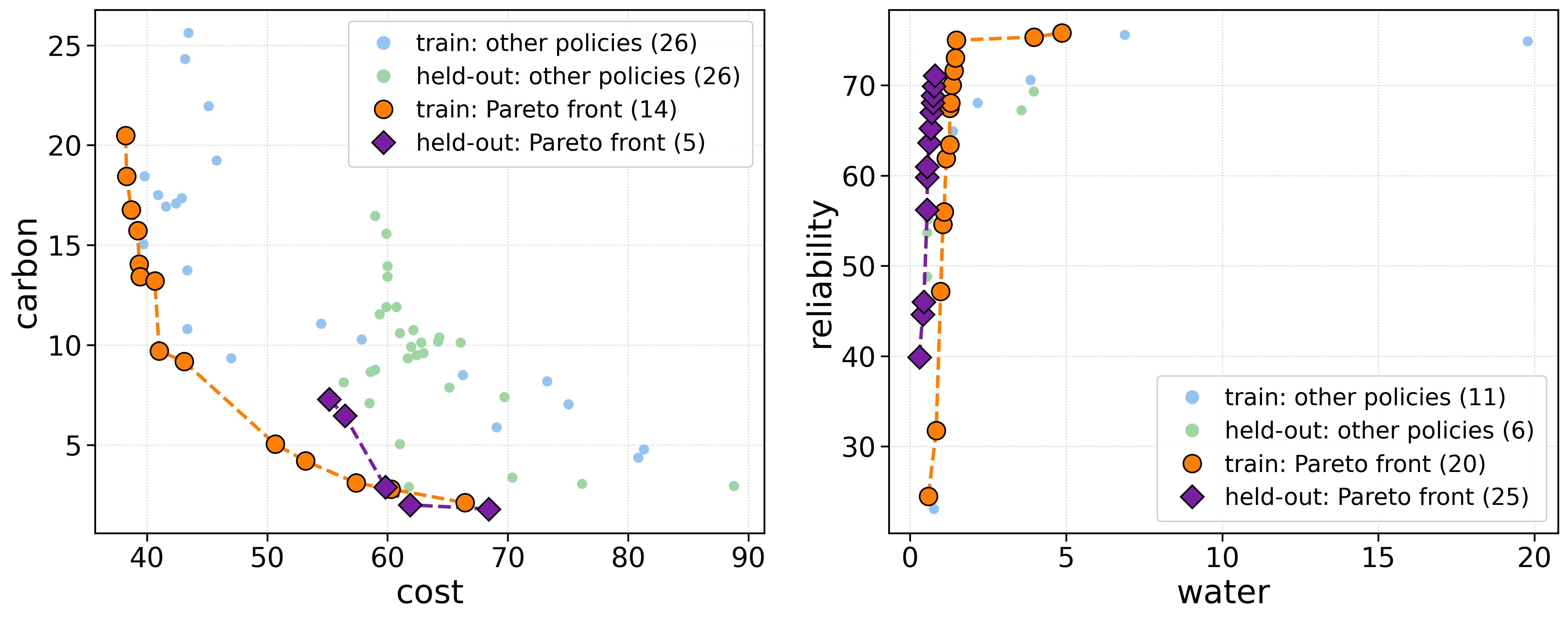}}\par}
\nopagebreak\vspace{4pt}
\caption{Recovered fronts. Cost $\times$ carbon is on the left half of each
panel and water $\times$ reliability on the right half. Fronts are split into
training and held-out site/scenario pairs. The reversible frontier is wider
but more clustered. The irreversible frontier is narrower but more evenly
spread, matching the sparsity gap in Table~\ref{tab:main}. The commit-once
frontier (single representative seed) spans a range comparable to the
reversible regime, with no post-day-zero retrofit to shape it. All flat-agent
fronts and all learning curves are in Appendix~\ref{app:figures}.}
\label{fig:fronts}
\endgroup
\vspace{0.5em}

\paragraph{Reversibility trades frontier width for coverage and
consistency:} The reversible regime reaches a wider raw frontier. An
unconstrained switching policy can locally chase whichever objective the
current preference favors. But that width is unevenly distributed. Sparsity is
$16.80$--$25.43$ under reversibility, against $9.38$--$9.58$ under lock-in.
The lock-in frontier is therefore the one a planner can actually interpolate
across at inference time. The gap is starker on anticipation. Under lock-in,
policies leave evaporative cooling in $88.9\%$ of scored episodes, with zero
seed-to-seed variance. They almost always do so at the earliest legal window.
Under free reversibility the rate falls to $60.8\%\pm43.0$, and switch timing
is scattered across the horizon (Appendix~\ref{app:figures},
Fig.~\ref{fig:antic}). Under commit-once, anticipation is not a meaningful
metric. Any adaptation must be made once, at $t=0$, through the initial
architecture choice. The recovered anticipation rate is trivially zero, with
zero variance across seeds, for exactly that reason.

\paragraph{Preference conditioning is confirmed everywhere:} An eleven-point
preference sweep confirms the intended trade-off on both objective pairs in
all three regimes. Correlation magnitudes are at least $0.88$ on
cost$\times$carbon (Table~\ref{tab:steer}). Per-regime sweep figures and
extended discussion are in Appendix~\ref{app:steer}. The water$\times$reliability
trade-off is confirmed everywhere. It is more gradual and evenly steerable
under lock-in. It is closer to an on/off switch under reversibility and
commit-once. This is consistent with the lower sparsity already reported for
the irreversible regime.

\begin{table}[H]
\centering\footnotesize
\caption{Preference-steerability correlations from an eleven-point sweep.
\emph{Own} is the correlation between an objective's weight and its realized
value, where positive is expected. \emph{Other} is the correlation with the
competing objective, where negative is expected. All trade-offs are
confirmed.}
\label{tab:steer}
\begin{tabular}{@{}llccl@{}}
\toprule
Regime & Objective pair & Corr.\ own & Corr.\ other & Verdict \\
\midrule
\textsc{irreversible} & cost $\times$ carbon       & $+0.95$ & $-0.98$ & confirmed \\
\textsc{irreversible} & water $\times$ reliability & $+0.85$ & $-0.65$ & confirmed \\
\textsc{reversible}   & cost $\times$ carbon       & $+0.93$ & $-0.96$ & confirmed \\
\textsc{reversible}   & water $\times$ reliability & $+0.65$ & $-0.52$ & confirmed \\
\textsc{commit-once}  & cost $\times$ carbon       & $+0.88$ & $-0.99$ & confirmed \\
\textsc{commit-once}  & water $\times$ reliability & $+0.59$ & $-0.54$ & confirmed \\
\bottomrule
\end{tabular}
\end{table}

\paragraph{The counterfactual model is honest about where it struggles:}
Rolling-origin backtesting shows that the LCM beats linear drift by a wide
margin on all tracked state dimensions. It is roughly comparable to
persistence on slowly drifting dimensions (grid carbon, price, and workload
demand). But it is measurably \emph{outperformed} by persistence on water
stress. Measured water-stress drift is small under the current calibration for
most sites (Appendix~\ref{app:lcm}, Fig.~\ref{fig:wm}). We flag this rather
than hide it. The LCM should not be trusted for model-based planning on the
water axis. None of the policy results above depend on it, since it is a
diagnostic tool and not a training-time model.

\vspace{-2pt}
\section{Discussion and Broader Impact}
\label{sec:discussion}
\vspace{-1pt}
The marginal water, carbon, and energy footprint of the AI build-out is
determined largely at planning time. Region, power pathway, cooling
technology, and schedule all matter. Representing irreversibility changes
which policies are learned. Across three commitment regimes, from fully
reversible to a single day-zero choice, lock-in produces the most consistently
anticipatory and evenly steerable policy. This is precisely the behavior a
planner needs in order to act on a constraint years before it binds.
Limitations are discussed in Appendix~\ref{app:limitations}. They cover the
market-scale figures, the workload-energy proxy, the interconnection-queue
interpretation, and the absence of verified multi-decade ground truth. None of
our empirical claims depend on the precision of the motivating market figures.
We release the environment, agents, baselines, and evaluation harness to lower
the barrier for further work on anticipatory infrastructure planning.

Our work also connects to a growing effort to assess AI systems as planning
aids for this same build-out. \emph{Watts If?} \citep{chouhan2026wattsifbenchmarkcounterfactual}
introduces DCBench-440. This is a benchmark of generative and multiple-choice
questions on counterfactual reasoning about data center commitments. It finds
that language models fail at exhaustively enumerating the actions available at
a decision gate. This holds even for models that score above 90\% on
source-grounded multiple choice. That failure of option-tracking is the
natural counterpart to the irreversibility studied here. An assistant that
omits a feasible option at a gate cannot weigh what a commitment forecloses.
Furthermore, we plan to integrate the sequential decision-making aspect of the
AI build-out into DCBench. This will let us compare the performance of AI
models, including language-model planners, against the preference-conditioned
agents and baselines released with this paper.

\newpage
\bibliographystyle{plainnat}
\bibliography{references}

@inproceedings{acpbench,
  author    = {Kokel, Harsha and Katz, Michael and Srinivas, Kavitha and Sohrabi, Shirin},
  editor    = {Walsh, Toby and Shah, Julie and Kolter, Zico},
  title     = {{ACPBench}: Reasoning About Action, Change, and Planning},
  booktitle = {Proceedings of the AAAI Conference on Artificial Intelligence (AAAI)},
  volume    = {39},
  number    = {25},
  pages     = {26559--26568},
  year      = {2025},
  doi       = {10.1609/aaai.v39i25.34857},
  url       = {https://doi.org/10.1609/aaai.v39i25.34857}
}

@article{acpbenchhard,
  author    = {Kokel, Harsha and Katz, Michael and Srinivas, Kavitha and Sohrabi, Shirin},
  title     = {{ACPBench Hard}: Unrestrained Reasoning about Action, Change, and Planning},
  journal   = {arXiv preprint arXiv:2503.24378},
  year      = {2025},
  doi       = {10.48550/arXiv.2503.24378},
  url       = {https://doi.org/10.48550/arXiv.2503.24378}
}

@misc{mckinsey2025,
  author       = {{McKinsey \& Company}},
  title        = {The cost of compute: A \$7 trillion race to scale data centers},
  howpublished = {McKinsey Quarterly},
  month        = apr,
  year         = {2025},
  url          = {https://www.mckinsey.com/industries/technology-media-and-telecommunications/our-insights/the-cost-of-compute-a-7-trillion-dollar-race-to-scale-data-centers},
  note         = {Accessed August 2026}
}

@techreport{iea2025energyai,
  author      = {{International Energy Agency}},
  title       = {Energy and {AI}},
  institution = {IEA},
  address     = {Paris},
  month       = apr,
  year        = {2025},
  url         = {https://www.iea.org/reports/energy-and-ai},
  note        = {World Energy Outlook Special Report; accessed August 2026}
}

@techreport{lbnl2025queued,
  author      = {Rand, Joseph and Manderlink, Nick and Zhang, Steven and Talley, Chris and Gorman, Will and Wiser, Ryan H. and Seel, Joachim and Kemp, Julie Mulvaney and Jeong, Seongeun and Kahrl, Fredrich},
  title       = {Queued Up: 2025 Edition, Characteristics of Power Plants Seeking Transmission Interconnection As of the End of 2024},
  institution = {Lawrence Berkeley National Laboratory},
  month       = apr,
  year        = {2025},
  url         = {https://emp.lbl.gov/queues},
  note        = {Interconnection-queue microdata used for site lead-time calibration; accessed August 2026}
}

@misc{goldman2025power,
  author       = {{Goldman Sachs Research}},
  title        = {{AI} to Drive 165\% Increase in Data Center Power Demand by 2030},
  howpublished = {Goldman Sachs Insights},
  year         = {2025},
  url          = {https://www.goldmansachs.com/insights/articles/ai-to-drive-165-increase-in-data-center-power-demand-by-2030},
  note         = {Accessed August 2026}
}

@misc{cbre2025capex,
  author       = {{CBRE Investment Management}},
  title        = {Data Centers: Ain't No Mountain High Enough},
  howpublished = {CBRE Investment Management Insights},
  year         = {2025},
  url          = {https://www.cbreim.com/insights/articles/data-centers-aint-no-mountain-high-enough},
  note         = {Hyperscaler committed capital through 2027; accessed August 2026}
}

@misc{wri_aqueduct40,
  author       = {{World Resources Institute}},
  title        = {Aqueduct 4.0: Updated Decision-Relevant Global Water Risk Indicators},
  howpublished = {WRI Aqueduct},
  year         = {2023},
  url          = {https://www.wri.org/aqueduct},
  note         = {Baseline and future (SSP/RCP) basin water-stress data, used for per-site water-stress level and forward drift}
}

@misc{eia930,
  author       = {{U.S. Energy Information Administration}},
  title        = {Hourly Electric Grid Monitor (EIA-930) and Electricity Data {API}},
  howpublished = {EIA},
  year         = {2026},
  url          = {https://www.eia.gov/electricity/gridmonitor/},
  note         = {Balancing-authority generation by fuel type and industrial electricity price, used for per-site carbon intensity and price calibration}
}

@misc{lbnl_queued_data,
  author       = {Rand, Joseph and Manderlink, Nick and Zhang, Steven and Talley, Chris and Gorman, Will and Wiser, Ryan H. and Seel, Joachim and Kemp, Julie Mulvaney and Jeong, Seongeun and Kahrl, Fredrich},
  title        = {Interconnection Queue Data File (through 2024)},
  howpublished = {LBNL Electricity Markets \& Policy},
  year         = {2025},
  url          = {https://emp.lbl.gov/queues},
  note         = {Project-level queue data used for per-site, per-source procurement lead-time calibration}
}

@misc{westwater2025,
  author       = {{West Water Research}},
  title        = {Estimating Data Center Water Demand: New Analysis Predicts 170\% Surge in Data Center Water Use by 2030 Amid {AI} Expansion},
  howpublished = {Water Market Insider},
  year         = {2025},
  url          = {https://westwaterresearch.com/wp-content/uploads/2025/03/2025-Q2-Water-Market-Insider-Data-Centers.pdf},
  note         = {Accessed August 2026}
}

@misc{jll2026outlook,
  author       = {{JLL Research}},
  title        = {2026 Global Data Center Outlook},
  howpublished = {JLL Research},
  year         = {2026},
  url          = {https://www.jll.com/en-jp/insights/market-outlook/global-data-centers},
  note         = {Shell-and-core and all-in construction cost per {MW}; accessed August 2026}
}

@misc{nrel_pvwatts,
  author       = {{National Renewable Energy Laboratory}},
  title        = {{PVWatts} Calculator and {API}},
  howpublished = {NREL},
  year         = {2026},
  url          = {https://developer.nrel.gov/docs/solar/pvwatts/v8/},
  note         = {Site-level on-site renewable capacity factor and seasonal volatility}
}

@misc{openmeteo,
  author       = {{Open-Meteo}},
  title        = {Open-Meteo Weather Forecast {API}},
  howpublished = {Open-Meteo},
  year         = {2026},
  url          = {https://open-meteo.com/},
  note         = {Hourly temperature series used as a cooling-load driver}
}

@misc{semianalysis_inferencex,
  author       = {{SemiAnalysis}},
  title        = {{InferenceMAX} / {InferenceX} Inference Benchmark Dataset},
  howpublished = {SemiAnalysis},
  year         = {2026},
  url          = {https://semianalysis.com/},
  note         = {Device throughput and TDP telemetry used to derive a workload-efficiency (J/token) proxy signal}
}

@misc{artificialanalysis,
  author       = {{Artificial Analysis}},
  title        = {Artificial Analysis Data {API}},
  howpublished = {Artificial Analysis},
  year         = {2026},
  url          = {https://artificialanalysis.ai/},
  note         = {Independent LLM-serving price and throughput signal, used as corroboration for the workload-efficiency proxy}
}

@inproceedings{tesauro2006,
  author    = {Tesauro, Gerald and Jong, Nicholas K. and Das, Rajarshi and Bennani, Mohamed N.},
  title     = {A Hybrid Reinforcement Learning Approach to Autonomic Resource Allocation},
  booktitle = {Proceedings of the 2006 IEEE International Conference on Autonomic Computing (ICAC)},
  pages     = {65--73},
  year      = {2006},
  doi       = {10.1109/ICAC.2006.1662383},
  url       = {https://doi.org/10.1109/ICAC.2006.1662383}
}

@inproceedings{tesauro2007,
  author    = {Tesauro, Gerald and Das, Rajarshi and Chan, Hoi and Kephart, Jeffrey O. and Levine, David W. and Rawson, Freeman and Lefurgy, Charles},
  title     = {Managing Power Consumption and Performance of Computing Systems Using Reinforcement Learning},
  booktitle = {Advances in Neural Information Processing Systems (NeurIPS)},
  volume    = {20},
  pages     = {1497--1504},
  year      = {2007},
  url       = {https://proceedings.neurips.cc/paper/2007/hash/5810284da8284fa6e810931295f51d8b-Abstract.html}
}

@inproceedings{das2008,
  author    = {Das, Rajarshi and Kephart, Jeffrey O. and Lefurgy, Charles and Tesauro, Gerald and Levine, David W. and Chan, Hoi},
  title     = {Autonomic Multi-agent Management of Power and Performance in Data Centers},
  booktitle = {Proceedings of the 7th International Joint Conference on Autonomous Agents and Multiagent Systems (AAMAS): Industry Track},
  pages     = {107--114},
  year      = {2008},
  url       = {https://dl.acm.org/doi/10.5555/1402298.1402315}
}

@techreport{gao2014,
  author      = {Gao, Jim},
  title       = {Machine Learning Applications for Data Center Optimization},
  institution = {Google},
  type        = {White Paper},
  year        = {2014},
  url         = {https://static.googleusercontent.com/media/research.google.com/en//pubs/archive/42542.pdf}
}

@inproceedings{lazic2018,
  author    = {Lazic, Nevena and Boutilier, Craig and Lu, Tyler and Wong, Eehern and Roy, Binz and Ryu, MK and Imwalle, Greg},
  title     = {Data Center Cooling Using Model-Predictive Control},
  booktitle = {Advances in Neural Information Processing Systems (NeurIPS)},
  volume    = {31},
  pages     = {3814--3823},
  year      = {2018},
  url       = {https://proceedings.neurips.cc/paper/2018/hash/3b2c1f93d399996d934bb61dbeeed4e6-Abstract.html}
}

@article{roijers2013,
  author  = {Roijers, Diederik M. and Vamplew, Peter and Whiteson, Shimon and Dazeley, Richard},
  title   = {A Survey of Multi-Objective Sequential Decision-Making},
  journal = {Journal of Artificial Intelligence Research},
  volume  = {48},
  pages   = {67--113},
  year    = {2013},
  doi     = {10.1613/jair.3987},
  url     = {https://doi.org/10.1613/jair.3987}
}

@article{hayes2022,
  author  = {Hayes, Conor F. and R{\u{a}}dulescu, Roxana and Bargiacchi, Eugenio and K{\"a}llstr{\"o}m, Johan and Roijers, Diederik M. and Rzońca, Patrick and Brys, Tim and Now{\'e}, Ann and Steckelmacher, Denis and Mannion, Patrick and Vrancx, Peter and Mannor, Shie and Vamplew, Peter and Howley, Enda},
  title   = {A Practical Guide to Multi-Objective Reinforcement Learning and Planning},
  journal = {Autonomous Agents and Multi-Agent Systems},
  volume  = {36},
  number  = {1},
  pages   = {26},
  year    = {2022},
  doi     = {10.1007/s10458-022-09552-y},
  url     = {https://doi.org/10.1007/s10458-022-09552-y}
}

@inproceedings{yang2019,
  author    = {Yang, Runzhe and Sun, Xingyuan and Narasimhan, Karthik},
  title     = {A Generalized Algorithm for Multi-Objective Reinforcement Learning and Policy Adaptation},
  booktitle = {Advances in Neural Information Processing Systems (NeurIPS)},
  volume    = {32},
  pages     = {15610--15621},
  year      = {2019},
  url       = {https://proceedings.neurips.cc/paper/2019/hash/4b2e8878939ebefd9197022e381017ef-Abstract.html}
}

@inproceedings{abels2019,
  author    = {Abels, Axel and Roijers, Diederik M. and Lenaerts, Tom and Now{\'e}, Ann and Steckelmacher, Denis},
  title     = {Dynamic Weights in Multi-Objective Deep Reinforcement Learning},
  booktitle = {Proceedings of the 36th International Conference on Machine Learning (ICML)},
  pages     = {11--20},
  year      = {2019},
  url       = {https://proceedings.mlr.press/v97/abels19a.html}
}

@inproceedings{xu2020,
  author    = {Xu, Jie and Tian, Yunsheng and Ma, Pingchuan and Rus, Daniela and Sueda, Shinjiro and Matusik, Wojciech},
  title     = {Prediction-Guided Multi-Objective Reinforcement Learning for Continuous Robot Control},
  booktitle = {Proceedings of the 37th International Conference on Machine Learning (ICML)},
  pages     = {10607--10616},
  year      = {2020},
  url       = {https://proceedings.mlr.press/v119/xu20h.html}
}

@inproceedings{alegre2022,
  author    = {Alegre, Lucas N. and Bazzan, Ana L. C. and da Silva, Bruno C.},
  title     = {Optimistic Linear Support and Successor Features as a Basis for Optimal Policy Transfer},
  booktitle = {Proceedings of the 39th International Conference on Machine Learning (ICML)},
  pages     = {394--413},
  year      = {2022},
  url       = {https://proceedings.mlr.press/v162/alegre22a.html}
}

@article{zitzler1999,
  author  = {Zitzler, Eckart and Thiele, Lothar},
  title   = {Multiobjective Evolutionary Algorithms: A Comparative Case Study and the Strength Pareto Approach},
  journal = {IEEE Transactions on Evolutionary Computation},
  volume  = {3},
  number  = {4},
  pages   = {257--271},
  year    = {1999},
  doi     = {10.1109/4235.797969},
  url     = {https://doi.org/10.1109/4235.797969}
}

@article{sutton1999,
  author  = {Sutton, Richard S. and Precup, Doina and Singh, Satinder},
  title   = {Between {MDP}s and Semi-{MDP}s: A Framework for Temporal Abstraction in Reinforcement Learning},
  journal = {Artificial Intelligence},
  volume  = {112},
  number  = {1-2},
  pages   = {181--211},
  year    = {1999},
  doi     = {10.1016/S0004-3702(99)00052-1},
  url     = {https://doi.org/10.1016/S0004-3702(99)00052-1}
}

@article{dietterich2000,
  author  = {Dietterich, Thomas G.},
  title   = {Hierarchical Reinforcement Learning with the {MAXQ} Value Function Decomposition},
  journal = {Journal of Artificial Intelligence Research},
  volume  = {13},
  pages   = {227--303},
  year    = {2000},
  doi     = {10.1613/jair.639},
  url     = {https://doi.org/10.1613/jair.639}
}

@inproceedings{vezhnevets2017,
  author    = {Vezhnevets, Alexander Sasha and Osindero, Simon and Schaul, Tom and Heess, Nicolas and Jaderberg, Max and Silver, David and Kavukcuoglu, Koray},
  title     = {{FeUdal} Networks for Hierarchical Reinforcement Learning},
  booktitle = {Proceedings of the 34th International Conference on Machine Learning (ICML)},
  pages     = {3540--3549},
  year      = {2017},
  url       = {https://proceedings.mlr.press/v70/vezhnevets17a.html}
}

@inproceedings{nachum2018,
  author    = {Nachum, Ofir and Gu, Shixiang and Lee, Honglak and Levine, Sergey},
  title     = {Data-Efficient Hierarchical Reinforcement Learning},
  booktitle = {Advances in Neural Information Processing Systems (NeurIPS)},
  volume    = {31},
  pages     = {3303--3313},
  year      = {2018},
  url       = {https://proceedings.neurips.cc/paper/2018/hash/e6384711491713d29bc63fc5eeb5ba4f-Abstract.html}
}

@article{nilim2005,
  author  = {Nilim, Arnab and El Ghaoui, Laurent},
  title   = {Robust Control of {Markov} Decision Processes with Uncertain Transition Matrices},
  journal = {Operations Research},
  volume  = {53},
  number  = {5},
  pages   = {780--798},
  year    = {2005},
  doi     = {10.1287/opre.1050.0216},
  url     = {https://doi.org/10.1287/opre.1050.0216}
}

@article{iyengar2005,
  author  = {Iyengar, Garud N.},
  title   = {Robust Dynamic Programming},
  journal = {Mathematics of Operations Research},
  volume  = {30},
  number  = {2},
  pages   = {257--280},
  year    = {2005},
  doi     = {10.1287/moor.1040.0129},
  url     = {https://doi.org/10.1287/moor.1040.0129}
}

@inproceedings{finn2017,
  author    = {Finn, Chelsea and Abbeel, Pieter and Levine, Sergey},
  title     = {Model-Agnostic Meta-Learning for Fast Adaptation of Deep Networks},
  booktitle = {Proceedings of the 34th International Conference on Machine Learning (ICML)},
  pages     = {1126--1135},
  year      = {2017},
  url       = {https://proceedings.mlr.press/v70/finn17a.html}
}

@article{das2026methane,
  title         = {Methane First: Sequencing Decarbonization Investment in the AI Data Center Boom},
  author        = {Das, Arun and Kar, Satyapragnya and Das, Rajarshi and
                   Chouhan, Pankaj and Panigrahi, Prabhas},
  year          = {2026},
  month         = oct,
  note          = {Available at SSRN},
  doi           = {10.2139/ssrn.7612338},
  url           = {https://ssrn.com/abstract=7612338}
}

@misc{chouhan2026wattsifbenchmarkcounterfactual,
      title={Watts If? A Benchmark for Counterfactual Reasoning About the AI Data Center Buildout}, 
      author={Pankaj Chouhan and Rajarshi Das and Aditya Dubey and Manminder Tomar and Sreekar Reddy and Samta Sharma and Satyapragnya Kar and PR Panigrahi},
      year={2026},
      eprint={2610.21397},
      archivePrefix={arXiv},
      primaryClass={cs.AI},
      url={https://arxiv.org/abs/2610.21397}, 
}

\newpage
\appendix
\section*{Appendix}

\section{Related Work}
\label{app:related}
\paragraph{RL for computing and energy systems:} RL has an established
record in autonomic power and performance management
\citep{tesauro2007,das2008,tesauro2006}. It has also been used for production
data center cooling control \citep{gao2014,lazic2018}. These works target
operational control at minute-to-hour granularity under a scalar objective. We
differ in horizon and objective structure. We study multi-year commitment
planning under four simultaneously active objectives. The binding difficulty
is irreversibility rather than tracking accuracy.

Closest to this paper is our companion study \emph{Methane First}
\citep{das2026methane}. It applies RL to the decarbonization of on-site
gas generation for data centers. It poses a scenario analysis on 69 U.S. data
centers with on-site gas turbines, growing to 350 by 2035. It asks how a fixed
annual budget should be split between methane leak detection and repair
(LDAR) and carbon capture and storage (CCS). It also asks whether the answer
depends on the global-warming-potential (GWP) window used to combine methane
and CO$_2$. That problem has a one-dimensional state, so it can be solved
exactly by dynamic programming. This makes it a rare setting in which a
model-free policy can be checked against a known optimum. A REINFORCE policy
trained over ten seeds recovers the exact plan in 17 of 20 runs of the
yearly-upkeep variant, and the rest are within 0.6\%. It recovers the exact
plan in every run of the one-time-purchase variant. Both the exact and learned
plans fund LDAR first under GWP20 and GWP100.

We view that result as supporting evidence for using RL where exact methods
stop scaling. This is the situation in the present paper. Four objectives,
gated commitments, lead times, and a manager and worker hierarchy produce a
state and action space far beyond exact dynamic programming. The two studies
are complementary in scope. \emph{Methane First} is a single-technology-pair
allocation with a verified optimum. This paper treats multi-year,
multi-objective build-out planning under irreversibility. There, evaluation
must instead rely on frontier quality, steerability, and held-out
generalization.

\paragraph{Multi-objective RL:} We adopt the utility-based view of
multi-objective sequential decision making \citep{roijers2013,hayes2022}. We
build on preference-conditioned value methods that recover a frontier with a
single network \citep{yang2019,abels2019}. Stronger frontier-coverage methods
\citep{xu2020,alegre2022} are natural extensions. We report the hypervolume
indicator \citep{zitzler1999} as the primary frontier-quality metric.

\paragraph{Hierarchical RL:} Our manager and worker decomposition draws on
temporal abstraction and the options framework
\citep{sutton1999,dietterich2000}. It also draws on manager and worker
architectures \citep{vezhnevets2017,nachum2018}. The manager is trained as a
semi-MDP over its commitment interval. Unlike goal-conditioned formulations,
it couples to the worker by restricting the worker's feasible action set to
energized capacity. It does not emit a target.

\paragraph{Robust and non-stationary RL:} Multi-decade horizons imply
drifting transition dynamics. This motivates robust MDP formulations
\citep{nilim2005,iyengar2005} and adaptation-oriented methods
\citep{finn2017}. We approximate this by training across a distribution of
site and scenario trajectories. We then evaluate on held-out combinations.

\paragraph{Planning benchmarks:} Our tiered, action-dependent MDP is closer in
spirit to structured, reasoning-heavy planning benchmarks such as ACPBench
\citep{acpbench,acpbenchhard} than to conventional continuous control RL
suites. Legality (the action mask) and multi-step consequence (pending
commitments and lead times) are first-class parts of the state. They are not
incidental physics. \emph{Watts If?} \citep{chouhan2026wattsifbenchmarkcounterfactual} applies
ACPBench Hard's question types to data center deal archetypes. It shows that
language models struggle most with enumerating legal actions. This is the same
notion of legality that our action masks make explicit for an RL agent.

\section{Extended Method: The Learned Counterfactual Model (LCM)}
\label{app:lcm}
The learned counterfactual model (LCM) is a $k{=}5$-network deep ensemble.
Each network outputs a Gaussian next-state delta $(\mu,\log\sigma^2)$ with a
clamped log-variance. It also has a deterministic reward-vector head. Training
uses negative log-likelihood plus mean-squared error on bootstrapped
minibatches. Predictive variance combines the mean of the per-network
variances with the variance of the per-network means. Epistemic and aleatoric
uncertainty are therefore both represented.

We validate with rolling-origin backtesting. We start from many points along
held-out trajectories. From each point, we roll the LCM forward under a fixed
random action sequence for up to 20 steps. We sample 16 paths per start to
propagate a predictive band. We then compare mean absolute error and $1\sigma$
empirical coverage against persistence and linear-drift baselines. The
comparison covers all 24 state dimensions. Ten are exogenous, such as carbon
intensity, price, water stress, and workload demand. Fourteen are endogenous,
such as installed cooling, energized capacity per source, and
pending-commitment countdowns.

\begin{figure}[H]
\centering
\includegraphics[width=\textwidth]{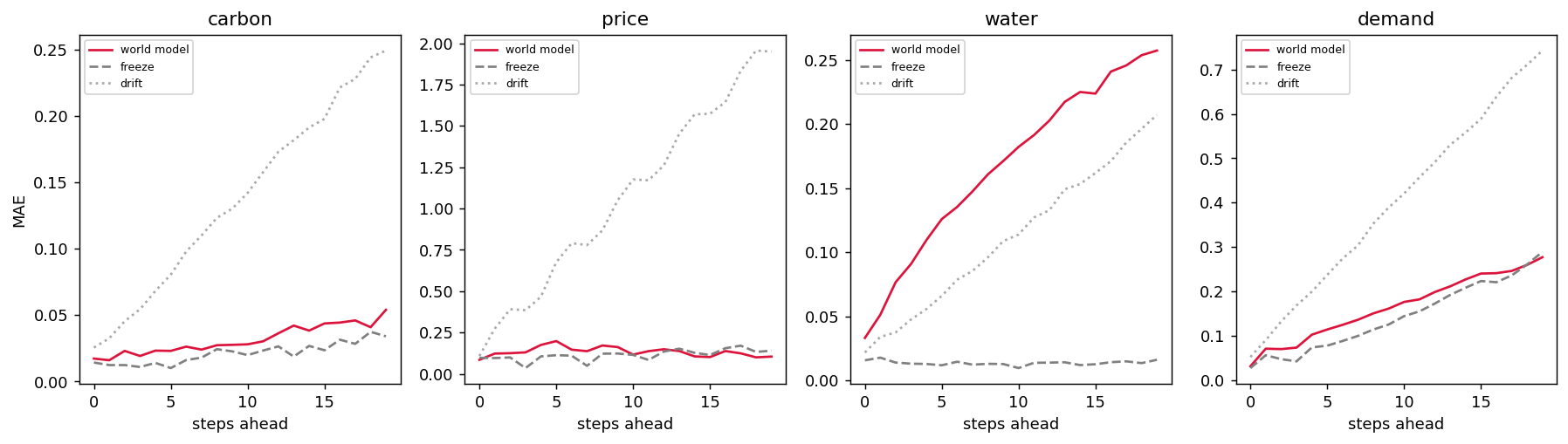}
\caption{LCM rolling-origin mean absolute error versus two baselines, by
steps ahead, on four held-out state dimensions. The LCM dominates linear
drift throughout. It is roughly on par with persistence on carbon, price, and
demand. It is measurably worse than persistence on water stress, an honest
limitation discussed in Section~\ref{sec:results}.}
\label{fig:wm}
\end{figure}

Figure~\ref{fig:wm} reports four representative dimensions. Persistence beats
the LCM on water stress specifically. This is consistent with the calibrated
water-stress growth rate being very small for most sites
(Appendix~\ref{app:data}). Freezing the last observed value is then close to
optimal by construction. A learned rollout model, in contrast, accumulates
compounding error. We treat this as a hard constraint on use, not a caveat to
be minimized. The model's own design note in the released code says that it
should not be used for model-based planning on any dimension where it does not
clearly beat both baselines.

\paragraph{Reading the backtest:} Figure~\ref{fig:wm} plots mean absolute
error against the number of steps ahead. It shows the LCM and the two
baselines on four held-out state dimensions. These are grid carbon intensity,
electricity price, workload demand, and water stress. Three patterns matter.
First, the linear-drift baseline is the weakest reference throughout.
Extrapolating a locally fitted slope compounds quickly over a multi-step
rollout. The LCM stays well below it on every plotted dimension at every
horizon. Second, on carbon, price, and demand, the LCM and persistence are
roughly on par. These are slowly drifting, largely exogenous signals. For such
signals, ``tomorrow looks like today'' is a strong predictor. Matching
persistence there is the realistic bar rather than a shortfall. It also
indicates that the LCM has not introduced spurious dynamics. Third, on water
stress, persistence is measurably better. We read this as a statement about
the signal and not only about the model. When the true per-site drift is small
relative to the noise a learned model must absorb, any residual error in the
predicted delta is added at each rolled-out step. Persistence incurs error only
through the drift itself.

\paragraph{Why the distinction matters for use:} The practical consequence is
a per-dimension trust map. The LCM can be queried for ``what if'' questions
about slow-moving price, carbon, and demand trajectories. There it is
competitive with the simplest sensible forecast. It also offers a benefit that
persistence cannot, which is conditioning on a counterfactual action sequence.
It cannot yet be used to justify a water-related commitment. Water stress is
the very axis on which the anticipation results of Section~\ref{sec:results}
are defined. This is why we keep the LCM strictly outside the training loop.
Policy learning uses the calibrated environment dynamics directly. None of the
frontier, steerability, or anticipation findings therefore inherit the LCM's
water-axis weakness. Two directions look promising. One is to model water
stress as a residual on top of persistence. The learned component would then
only have to explain departures from a frozen state. The other is to
recalibrate on scenarios with materially higher water-stress growth. There the
signal-to-noise ratio favors a learned model.

\section{Preference Steerability}
\label{app:steer}
Table~\ref{tab:steer} (Section~\ref{sec:results}) reports steerability
correlations from an eleven-point preference sweep. \emph{Own} is the
correlation between an objective's weight and its realized value, where
positive is expected. \emph{Other} is the correlation with the competing
objective, where negative is expected. The cost$\times$carbon trade-off is
close to linear in all three regimes, with magnitudes of at least $0.88$. The
water$\times$reliability trade-off is confirmed everywhere, but it takes a
different shape under lock-in. The irreversible regime shows a gradual,
near-linear decline in reliability as weight shifts toward water across the
full preference range. The reversible and commit-once regimes are close to
flat over most of the range. They drop sharply only at the extreme weight on
water (Fig.~\ref{fig:steer_all}). In practical terms, the lock-in frontier
gives a planner a usable, continuously adjustable water-reliability dial. The
other two regimes behave closer to an on-off switch.

\paragraph{Cost $\times$ carbon:} On this pair, the three regimes are nearly
indistinguishable in the quality of steering. The own-objective correlations
are 0.95, 0.93, and 0.88. The cross-objective correlations range from $-0.96$
to $-0.99$. All of them are strong. Shifting weight between cost and carbon
therefore moves the realized outcomes along the trade-off in the intended
direction, almost monotonically. Commit-once has the lowest own-correlation
(0.88) and the strongest cross-correlation ($-0.99$). This suggests that, with
a single day-zero choice, realized carbon is very tightly tied to the
preference. Realized cost responds a little less cleanly. A plausible reason
is that the one-time architecture and source choice is discrete, and each
choice carries its own cost. This is an interpretation of the correlations
and not a tested mechanism.

\paragraph{Water $\times$ reliability:} Here the regimes separate. Under
\textsc{irreversible}, both correlations are high (own $+0.85$, other
$-0.65$). Under \textsc{reversible} (own $+0.65$, other $-0.52$) and
\textsc{commit-once} (own $+0.59$, other $-0.54$), they are noticeably weaker.
These values are consistent with the sweep figures. A near-linear decline is
well described by a linear correlation. A flat-then-cliff curve is not. The
lower values under reversibility and commit-once therefore partly reflect the
\emph{shape} of the response. They do not necessarily indicate a failure of
conditioning. The trade-off is still confirmed in every regime. What differs
is how much of the preference range is useful for steering.

\paragraph{Why lock-in may help the dial:} One plausible reading is consistent
with the sparsity results of Table~\ref{tab:main}. When cooling can only
change at spaced windows, the agent must hedge its water-related commitment
against the whole horizon. This produces a response that varies smoothly with
the water weight. When the agent can switch freely, or choose once, the
cooling decision collapses into a nearly binary choice. The choice is made
only when the water weight becomes large enough to outweigh the alternative.
This reading is consistent with our experiments but not proved by them.
Isolating it would require probing the cooling-architecture decision directly
across the sweep.

\paragraph{Caveats:} The sweep uses eleven preference points and a single
representative run per regime. Correlations summarize monotone association.
They do not capture slope or the size of the achievable range. They should be
read alongside the sweep figures and the hypervolume and sparsity results, not
in isolation.

\begingroup\captionsetup{type=figure}
{\centering\subcaptionbox{\textsc{irreversible}.}{\includegraphics[width=0.9\textwidth]{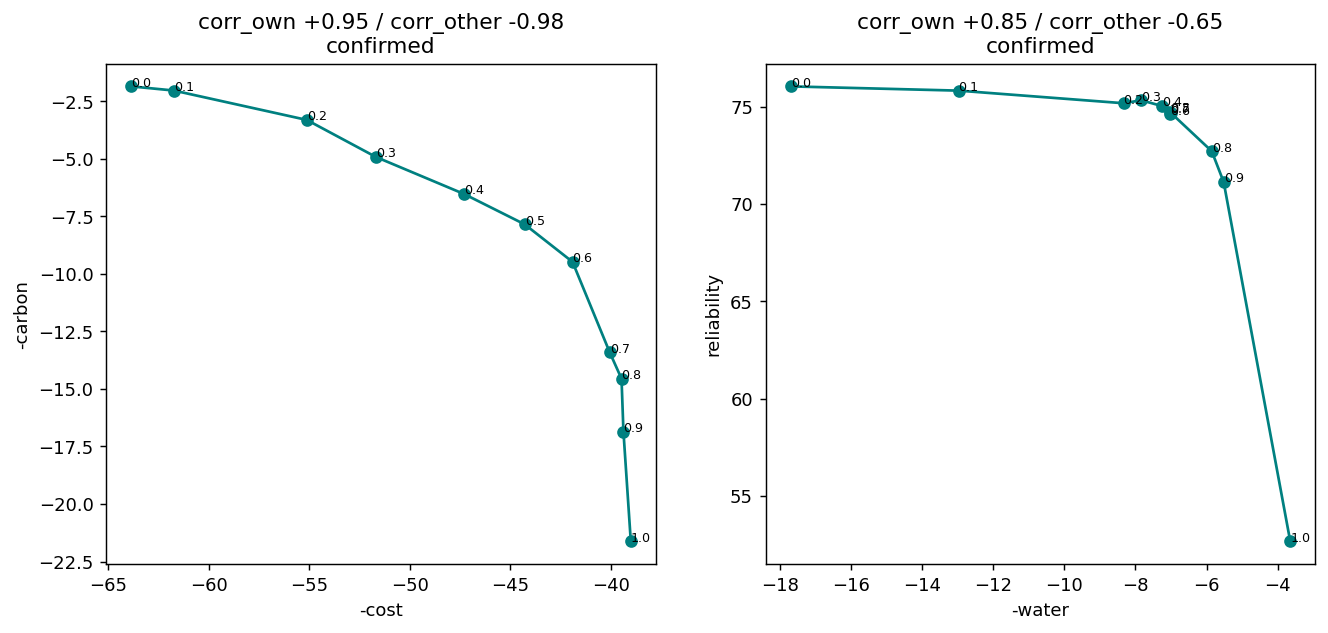}}\par}
\vspace{0.5em}
{\centering\subcaptionbox{\textsc{reversible}.}{\includegraphics[width=0.9\textwidth]{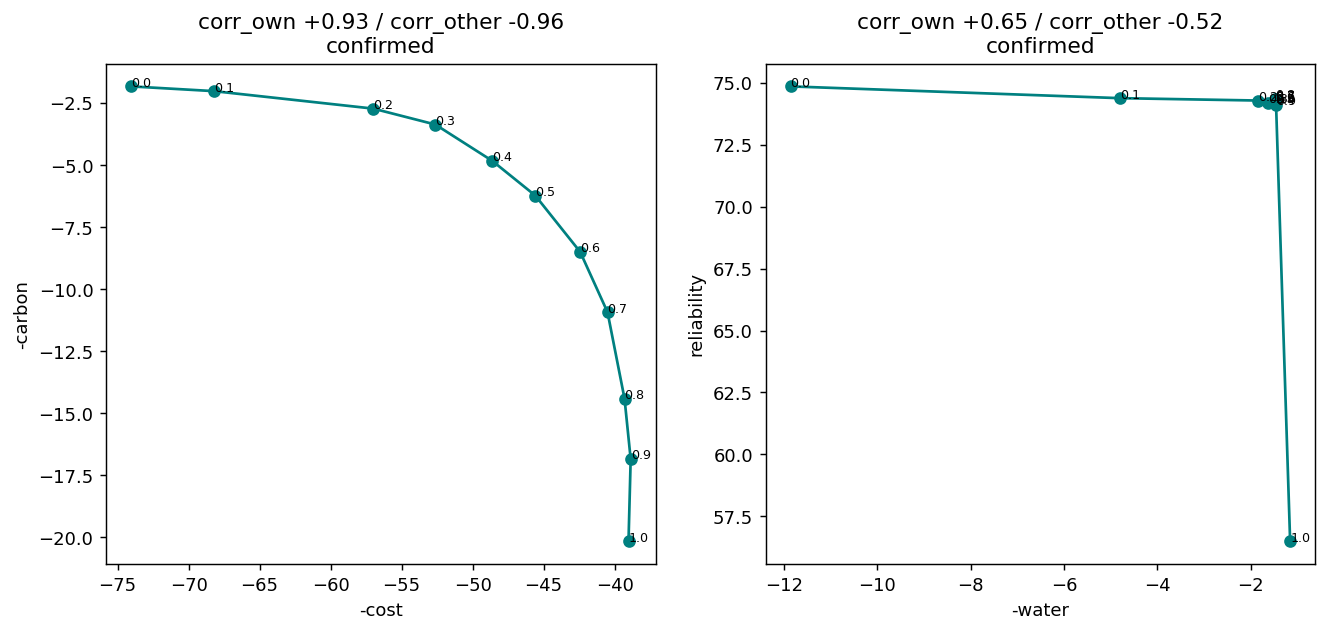}}\par}
\vspace{0.5em}
{\centering\subcaptionbox{\textsc{commit-once}.}{\includegraphics[width=0.9\textwidth]{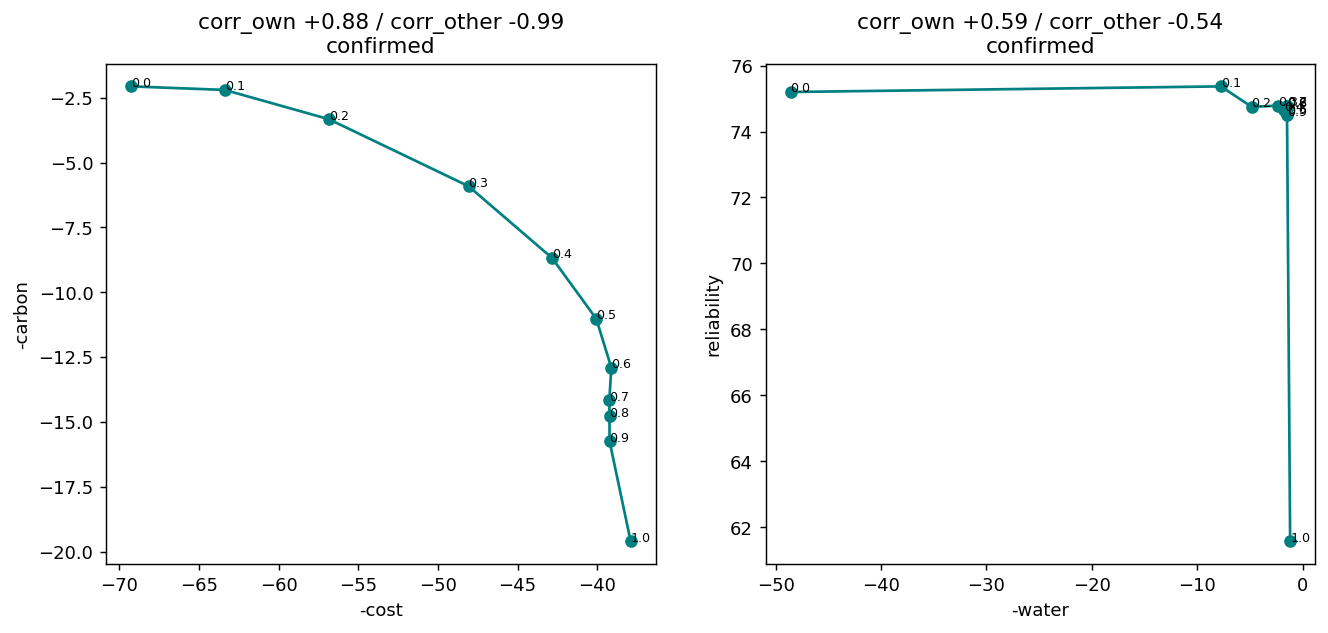}}\par}
\nopagebreak\vspace{4pt}
\caption{Preference sweeps underlying Table~\ref{tab:steer}. The left half of
each panel shows cost $\times$ carbon. The right half shows water $\times$
reliability. Note the smoother, more gradual slope on the water axis under
lock-in, against the flatter-then-cliff shape in the other two regimes.}
\label{fig:steer_all}
\endgroup
\vspace{0.5em}

\section{Extended Results}
\label{app:figures}
This section expands on the headline results of Section~\ref{sec:results}. We
first read Table~\ref{tab:main} more closely. We then discuss the flat-agent
and commit-once fronts, the anticipation figure, and the learning curves. All
figures here show a single representative seed per configuration.
Table~\ref{tab:main} aggregates three seeds. The figures therefore illustrate
the shape of the results, and the table carries the quantitative claims.

\subsection{Reading the hypervolume table more closely}

\paragraph{In-distribution versus held-out:} The most striking feature of
Table~\ref{tab:main} is the gap between in-distribution and held-out
cost$\times$carbon hypervolume. In-distribution values are roughly
$209$--$251$. Held-out values are $10$--$47$ across all six configurations.
Water$\times$reliability hypervolume degrades far less. It goes from
$1295$--$1323$ in-distribution to $931$--$1099$ held-out. The held-out set
contains two unseen sites and one unseen scenario. The cost$\times$carbon
plane is evidently more sensitive to the site-specific price and grid-carbon
levels that differ most across locations. Absolute held-out hypervolumes
should therefore be compared across agents within a regime. They should not be
compared across regimes or against the in-distribution column. We report this
gap openly because it bounds how far the learned frontiers can be trusted at
unseen locations.

\paragraph{Where hierarchy helps most on held-out data:} The clearest
hierarchical advantage on held-out data is on water$\times$reliability under
lock-in. The hierarchical agent reaches $1088.1\pm47.2$ and the flat agent
reaches $930.6\pm24.3$. This is an improvement of roughly 17\%, well outside
the reported spread. Under reversibility, the two agents are close on this
held-out view ($1091.2$ vs.\ $1095.8$). Under commit-once, the hierarchical
mean is higher ($1098.9$ vs.\ $1023.3$). However, the flat agent's standard
deviation ($100.6$) is large enough that we do not claim a difference. On
held-out cost$\times$carbon, the differences are small relative to the
standard deviations in every regime. One example is $10.4\pm8.6$ vs.\
$14.0\pm10.3$ under lock-in. We regard the two agents as comparable there.

\paragraph{Variance across seeds:} A recurring pattern is that the
hierarchical agent has the smaller standard deviation on in-distribution
cost$\times$carbon hypervolume in all three regimes. The values are $6.7$ vs.\
$12.0$, $8.8$ vs.\ $19.2$, and $5.8$ vs.\ $12.9$. Sparsity has two exceptions.
Under commit-once, the hierarchical agent is both worse and far more variable
($34.68\pm21.70$). Under reversibility, the \emph{flat} agent is the volatile
one ($25.43\pm15.44$ against $16.80\pm2.02$). With only three seeds, standard
deviations are themselves noisy. We treat these as indications of stability
and not as precise estimates.

\subsection{Recovered fronts}

\paragraph{Flat-agent fronts under lock-in and free reversibility:}
Figure~\ref{fig:fronts_flat} is the flat-agent companion to
Figure~\ref{fig:fronts}. The qualitative contrast between the regimes survives
the removal of the hierarchy. The reversible frontier is the broader of the
two, and the irreversible frontier is the more compact. This matters for
interpretation. It indicates that the width-versus-coverage trade-off is
driven mainly by the commitment regime and not by the agent architecture. The
hierarchical agent mainly sharpens the result. It gives higher hypervolume
and, under reversibility, more consistent sparsity. It does not change the
character of the result.

\paragraph{Commit-once fronts for both agents:} Figure~\ref{fig:fronts_co}
places the hierarchical and flat agents side by side under commit-once. Both
recover a cost$\times$carbon and a water$\times$reliability trade-off with a
range comparable to the reversible regime. This is consistent with the
regime's structure. A single day-zero choice of site architecture and power
pathway carries none of the ongoing switching friction that narrows the
irreversible frontier. The policy is therefore free to spread across the
objectives through the initial selection alone. The flat agent is the more
evenly covered of the two in Table~\ref{tab:main} ($20.59\pm5.07$ against
$34.68\pm21.70$). The figure shows one seed, and the hierarchical sparsity has
a very large spread across seeds. A single illustrative front should
therefore not be over-read in either direction.

\begingroup\captionsetup{type=figure}
{\centering\subcaptionbox{\textsc{irreversible}, flat agent.}{\includegraphics[width=0.9\textwidth]{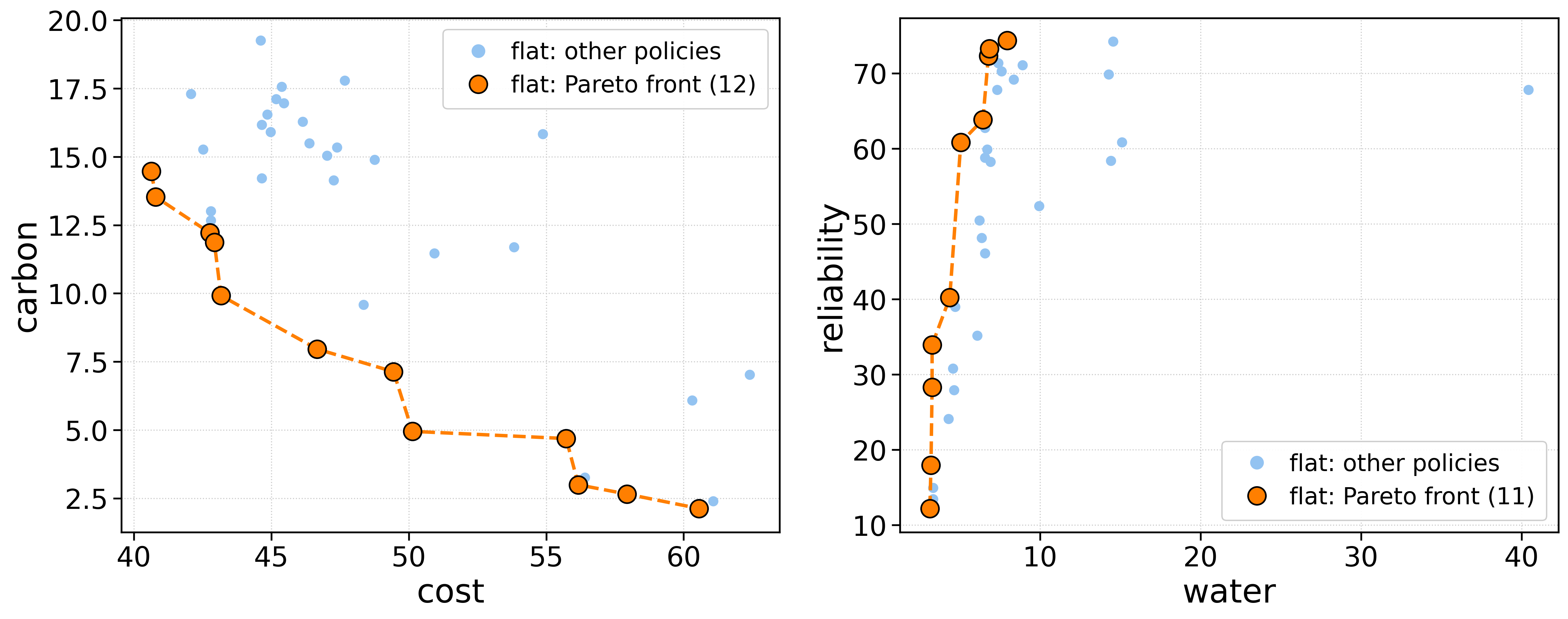}}\par}
\vspace{0.5em}
{\centering\subcaptionbox{\textsc{reversible}, flat agent.}{\includegraphics[width=0.9\textwidth]{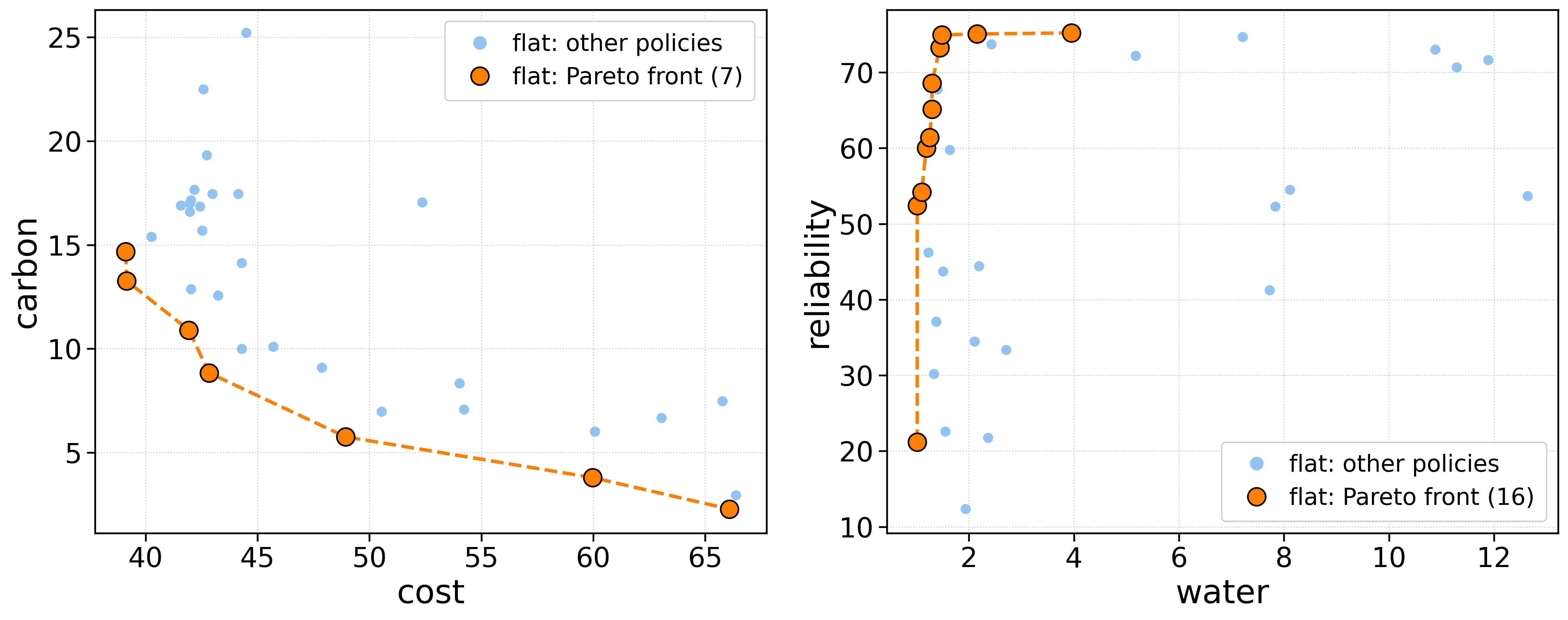}}\par}
\nopagebreak\vspace{4pt}
\caption{Recovered fronts for the flat (non-hierarchical) agent under
lock-in and free reversibility, companion to Figure~\ref{fig:fronts}.}
\label{fig:fronts_flat}
\endgroup
\vspace{0.5em}

\begingroup\captionsetup{type=figure}
{\centering\subcaptionbox{\textsc{commit-once}, hierarchical agent.}{\includegraphics[width=0.9\textwidth]{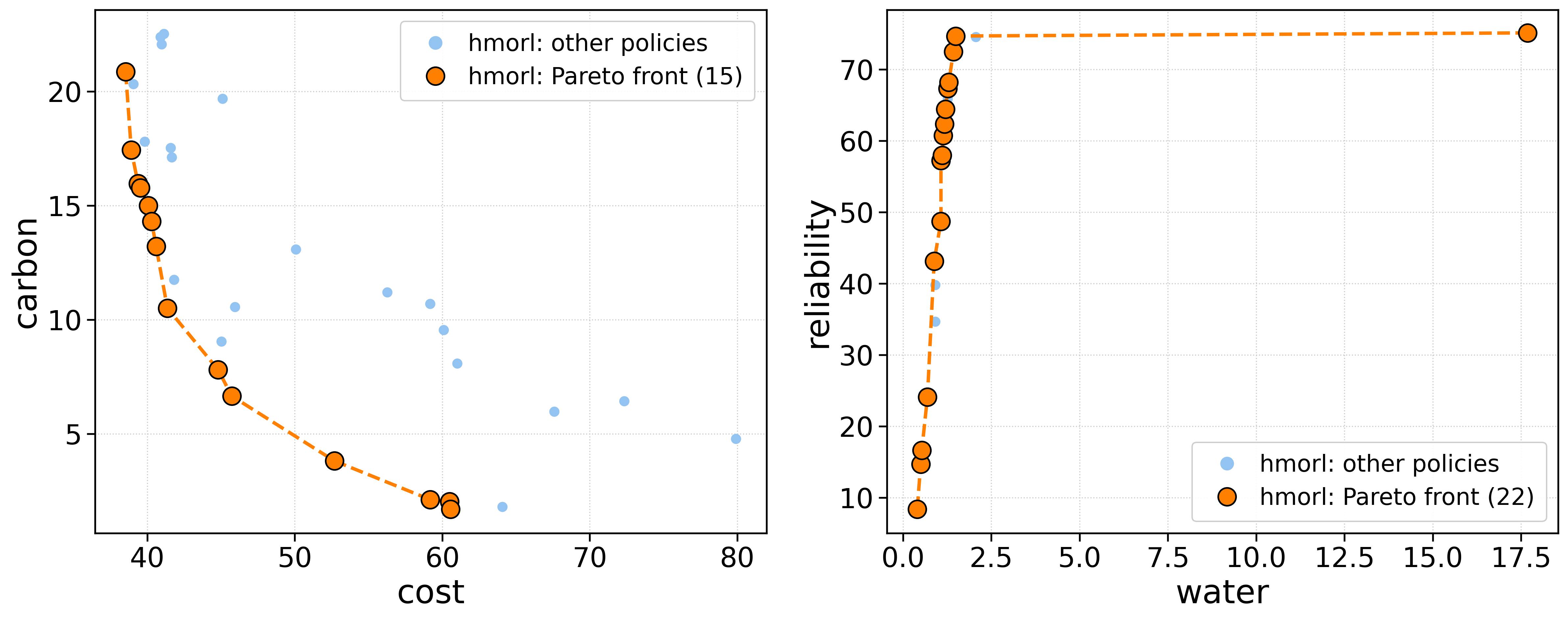}}\par}
\vspace{0.5em}
{\centering\subcaptionbox{\textsc{commit-once}, flat agent.}{\includegraphics[width=0.9\textwidth]{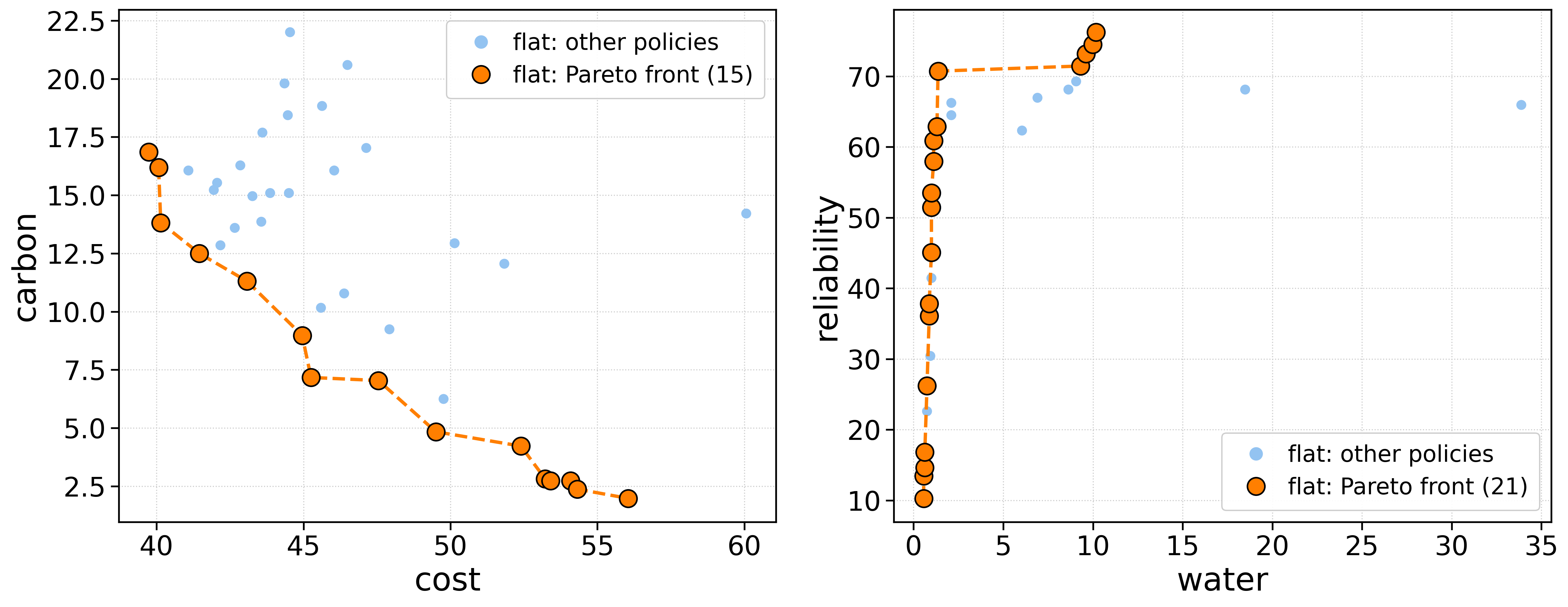}}\par}
\nopagebreak\vspace{4pt}
\caption{Recovered fronts under \textsc{commit-once} for the hierarchical
and flat agents, split into training and held-out site/scenario pairs. Each
panel shows a single representative seed. Multi-seed hypervolume and sparsity
aggregates are reported in Table~\ref{tab:main}. Both agents recover a
cost$\times$carbon and water$\times$reliability trade-off comparable in range
to the reversible regime. This is consistent with a single day-zero choice
carrying none of the ongoing switching friction that shapes the irreversible
regime. Note from Table~\ref{tab:main} that the flat agent's front is in fact
the more evenly covered of the two under this regime.}
\label{fig:fronts_co}
\endgroup
\vspace{0.5em}

\subsection{Anticipatory retrofit timing}

Figure~\ref{fig:antic} plots, for each regime, the number of episodes that
leave evaporative cooling in each year (bars). It overlays the water-stress
trajectory of a representative draw. Together with the anticipation rates in
Section~\ref{sec:results}, it shows the clearest behavioral difference between
the regimes.

\paragraph{Lock-in concentrates retrofits at the first legal window:} Under
\textsc{irreversible}, nearly all retrofits fall at the first available
window. This matches the $88.9\%$ anticipation rate with zero seed-to-seed
variance. The policies agree not only on whether to retrofit but on when. This
is what one expects when a window is rare. Waiting for the next one costs a
12-quarter delay. An agent that expects water stress to rise has little
incentive to defer, so the earliest window dominates. This is the anticipatory
behavior a planner wants, since the constraint binds years later.

\paragraph{Free reversibility scatters and weakens the signal:} Under
\textsc{reversible}, the few retrofits that occur are spread across the
horizon. The anticipation rate drops to $60.8\%\pm43.0$. A standard deviation
of 43 points on a rate bounded between 0 and 100 implies something specific.
Seeds disagree sharply about whether to leave evaporative cooling at all, and
not merely about timing. The intuition is that switching is cheap and always
available. There is then no cost to waiting. The learned policy can postpone
indefinitely and react to conditions as they emerge. This is precisely the
myopic behavior that static scoring implies.

\paragraph{Commit-once has no bars by construction:} With no post-day-zero
window, no retrofit can occur, and the panel shows none. Any adaptation to
future water stress must be expressed through the initial architecture choice.
This is why the anticipation metric is not defined for this regime.

\paragraph{On the reference line:} The dashed line at $0.85$ in the figure is
retained only for visual continuity with earlier internal reports. The
reported anticipation rate uses the calibrated, site-relative rise criterion of
Appendix~\ref{app:state}. The line should not be read as the scoring
threshold.

\begingroup\captionsetup{type=figure}
{\centering\subcaptionbox{\textsc{irreversible}.}{\includegraphics[width=0.48\textwidth]{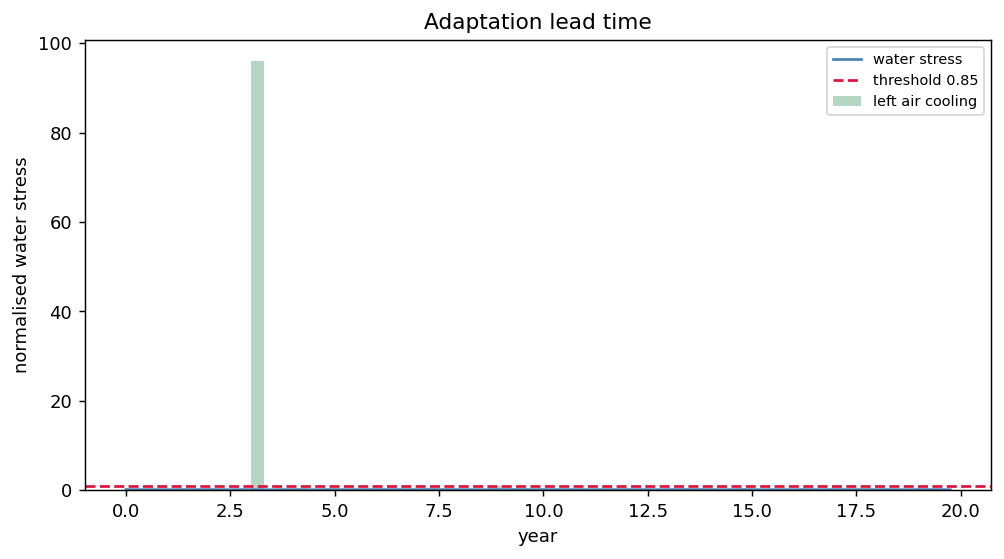}}\hfill
\subcaptionbox{\textsc{reversible}.}{\includegraphics[width=0.48\textwidth]{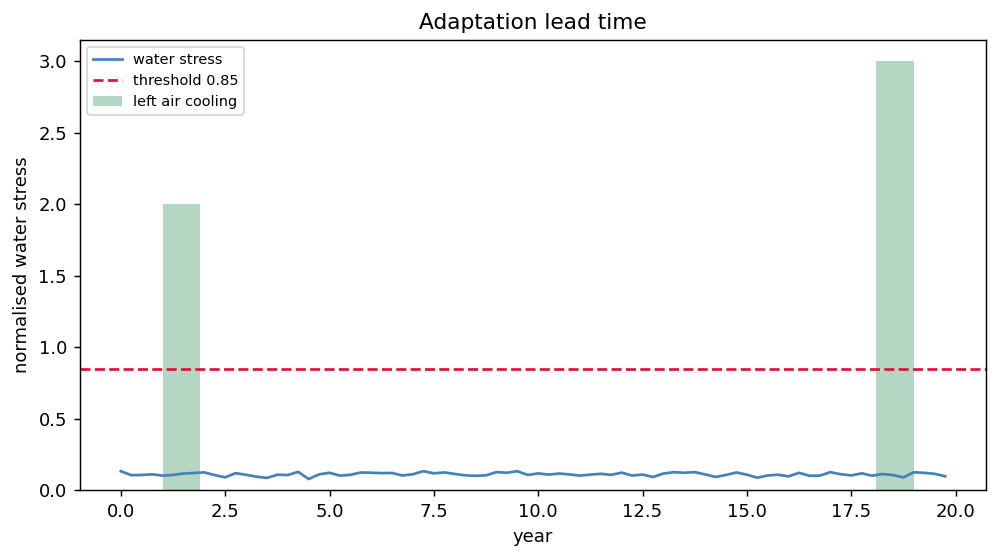}}\par}
\vspace{0.5em}
{\centering\subcaptionbox{\textsc{commit-once}.}{\includegraphics[width=0.48\textwidth]{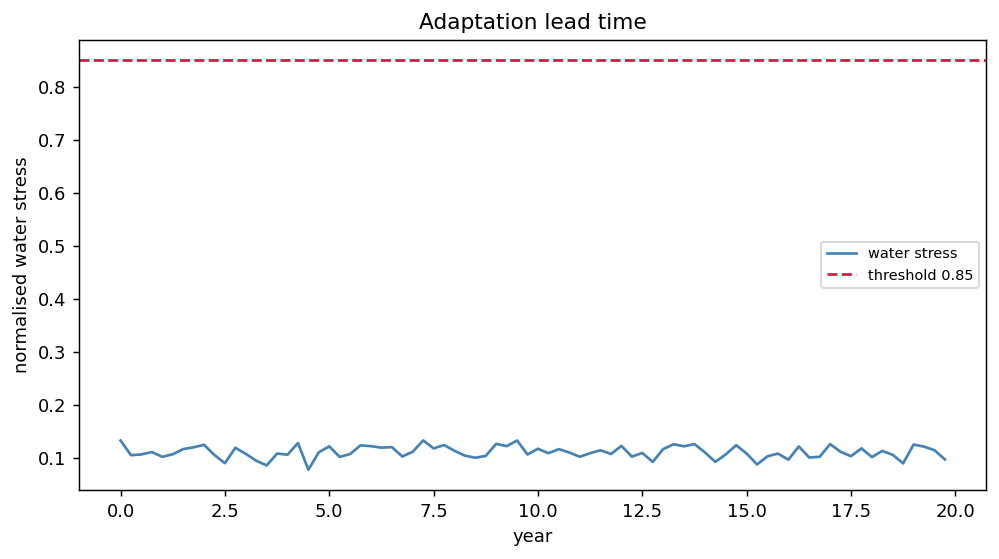}}\par}
\nopagebreak\vspace{4pt}
\caption{Retrofit timing (bars, count of episodes leaving evaporative
cooling in that year) against the water-stress trajectory of a
representative draw, for all three regimes. The dashed reference line at 0.85
is retained for visual comparability with earlier internal reports. The
anticipation rate actually reported in Table~\ref{tab:main} and
Section~\ref{sec:results} uses the calibrated, site-relative rise criterion
described in Appendix~\ref{app:state}, not this fixed line. Under lock-in,
switches concentrate almost entirely at the first legal window. Under free
reversibility they are few and scattered. Under commit-once there is no
post-day-zero window, so no bars appear at all.}
\label{fig:antic}
\endgroup
\vspace{0.5em}

\subsection{Learning dynamics}

Figure~\ref{fig:learning} shows three quantities for each regime and agent. They
are the scalarized return at a fixed preference vector and the two
hypervolume views over training.

\paragraph{Stability of the hierarchical agent:} Across the regimes, the
hierarchical curves are the steadier ones. This is consistent with the lower
seed-to-seed standard deviation of in-distribution hypervolume in
Table~\ref{tab:main}. The manager bootstraps over its four-quarter commitment
interval. Credit for a long-lived decision is therefore assigned directly and
not diffused over many single-quarter steps. This is a plausible source of the
smoother training signal.

\paragraph{Noise in the flat agent:} The flat agent chooses over the joint
$12\times 7{=}84$ action space with a single bootstrap interval. Its curves
are visibly noisier. This is most pronounced for the flat commit-once run,
which is noisier throughout than its hierarchical counterpart. The same
hierarchical-versus-flat stability gap appears under irreversible and
reversible. Yet on commit-once, the flat agent's final multi-seed sparsity is
the lower, more even, of the two. The learning curves and the final-frontier
metrics therefore tell different stories on this regime. Neither alone is a
sufficient summary.

\paragraph{How to read the three panels:} The scalarized-return panel tracks
progress on a single operating point. The two hypervolume panels measure the
frontier recovered across preferences. A curve can plateau on the first while
the hypervolume continues to change. Improving coverage at other preferences
does not alter return at the fixed one. We therefore read all three together.
We also treat late-training fluctuations in hypervolume as a reason to report
multi-seed means as in Table~\ref{tab:main}, and not the final checkpoint of
one run.

\begingroup\captionsetup{type=figure}
{\centering\subcaptionbox{\textsc{irreversible}, flat.}{\includegraphics[width=\textwidth]{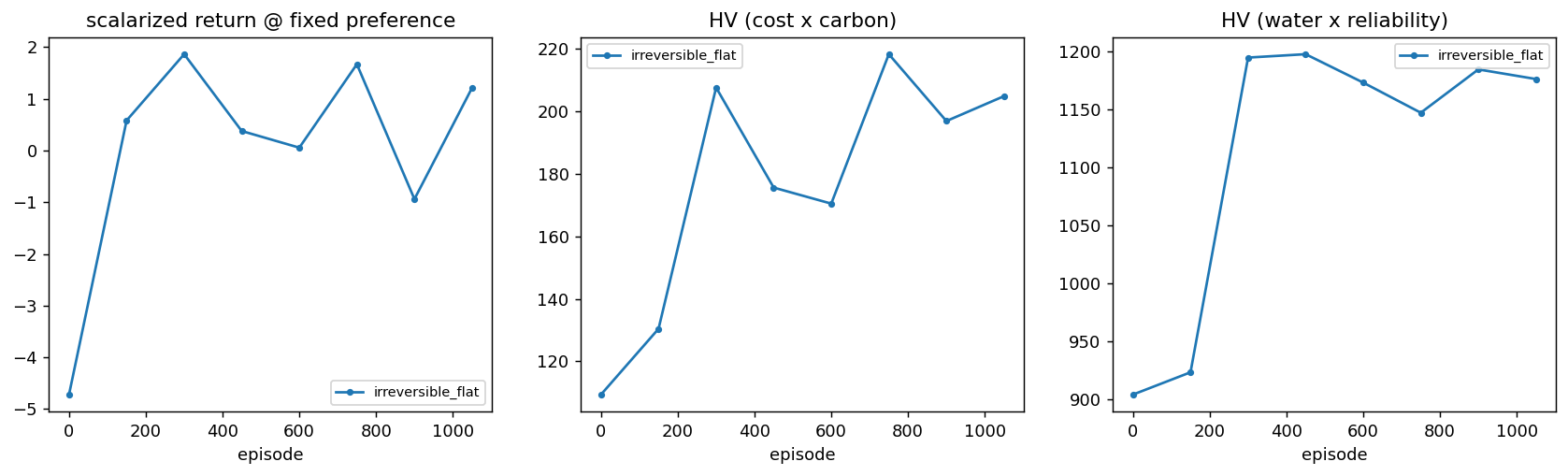}}\par}
\vspace{0.5em}
{\centering\subcaptionbox{\textsc{irreversible}, hierarchical.}{\includegraphics[width=\textwidth]{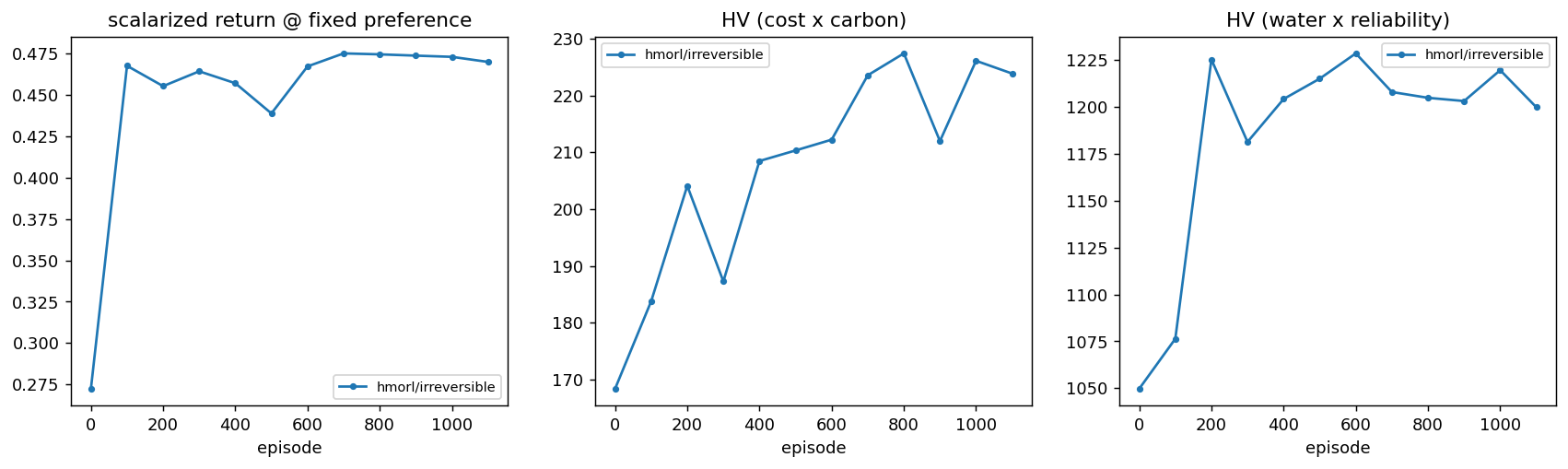}}\par}
\vspace{0.5em}
{\centering\subcaptionbox{\textsc{reversible}, flat.}{\includegraphics[width=\textwidth]{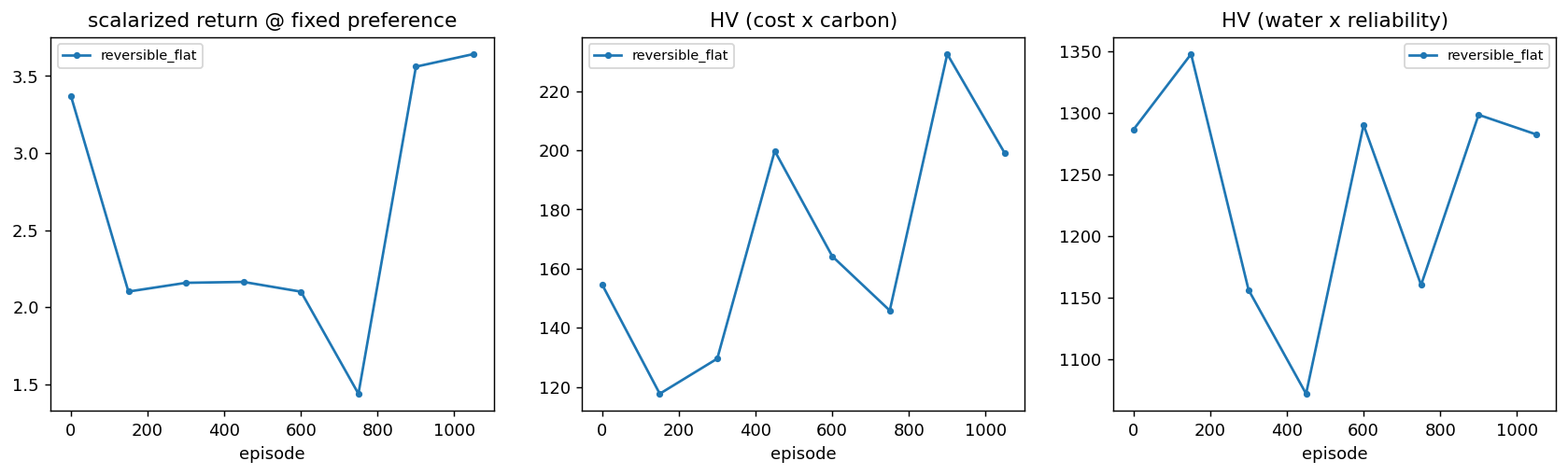}}\par}
\vspace{0.5em}
{\centering\subcaptionbox{\textsc{reversible}, hierarchical.}{\includegraphics[width=\textwidth]{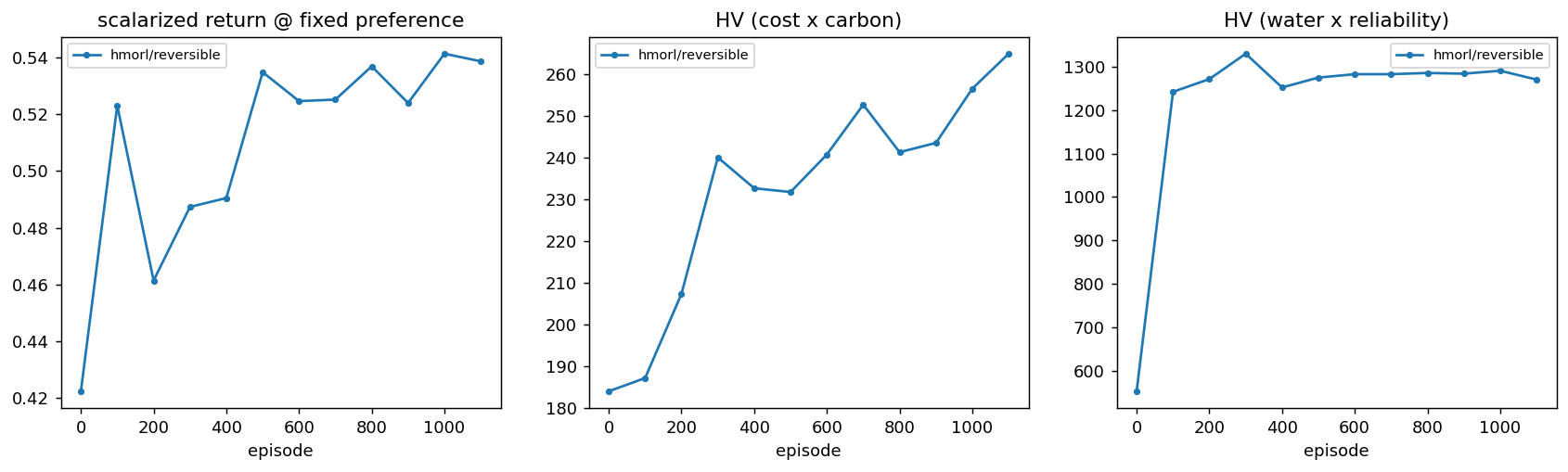}}\par}
\vspace{0.5em}
{\centering\subcaptionbox{\textsc{commit-once}, flat.}{\includegraphics[width=\textwidth]{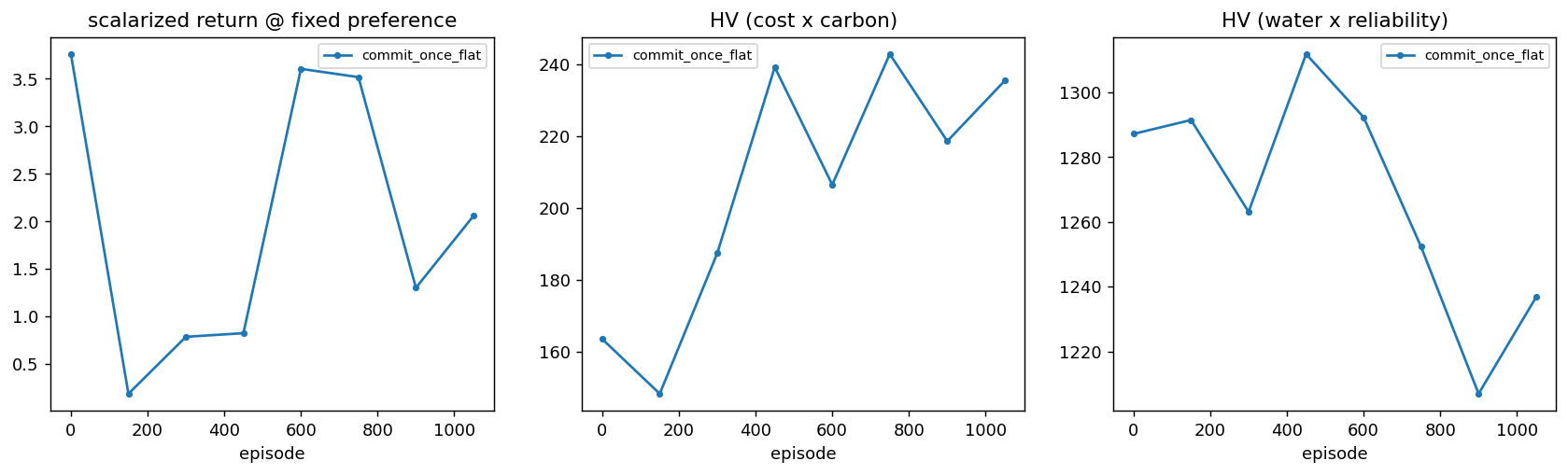}}\par}
\vspace{0.5em}
{\centering\subcaptionbox{\textsc{commit-once}, hierarchical.}{\includegraphics[width=\textwidth]{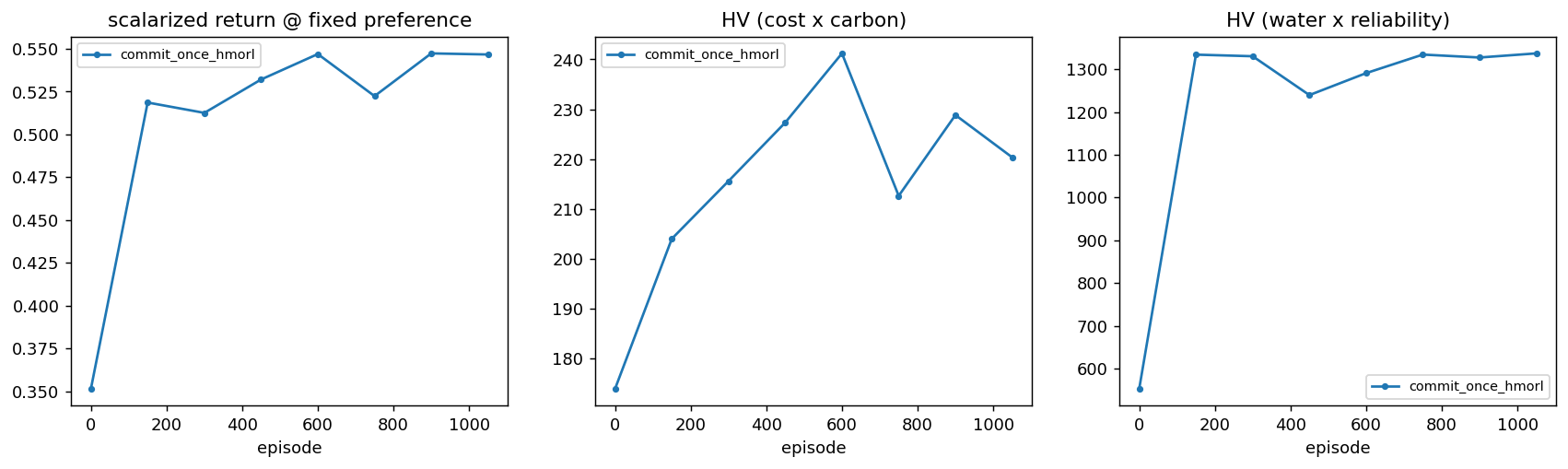}}\par}
\nopagebreak\vspace{4pt}
\caption{Scalarized return at a fixed preference vector, hypervolume (cost
$\times$ carbon), and hypervolume (water $\times$ reliability) over
training, for all three regimes and both agent types. Curves are shown for
a single representative seed each. Multi-seed hypervolume and sparsity
aggregates for all three regimes are reported in Table~\ref{tab:main}. The
flat \textsc{commit-once} run is visibly noisier throughout than the
hierarchical run. This echoes the same hierarchical-vs-flat stability gap seen
under \textsc{irreversible} and \textsc{reversible}. It holds even though, on
this regime, the flat agent's final multi-seed sparsity is the lower (more
even) of the two.}
\label{fig:learning}
\endgroup
\vspace{0.5em}

\section{Full State, Action, and Site Detail}
\label{app:state}
The 24-dimensional state comprises 10 exogenous signals and 14 endogenous
variables. The exogenous signals are grid carbon intensity and its peak,
electricity price and volatility, water stress, cooling load and its peak,
workload demand, interconnection-queue pressure, and on-site renewable
availability. The endogenous variables are the installed cooling one-hot,
energized capacity per source, pending-commitment countdowns, the
retrofit-window indicator, cumulative capital, and normalized elapsed time.

The manager chooses among 3 cooling architectures times 4 power sources,
giving 12 actions. The worker chooses among 7 dispatch mixes. These are single
sources, pairwise blends, and an equal four-way blend. Both are masked.
Outside a retrofit window, the manager's cooling component is pinned to the
installed architecture. The worker's mask excludes any mix drawing on
unenergized capacity.

The binding event for the anticipation metric is a water-stress level
materially above what was underwritten at day zero. This is a relative rise
past a per-site floor. It is not a crossing of a fixed global number. The
fixed 0.85 line shown in Figure~\ref{fig:antic} is kept only for visual
continuity with earlier internal reports. It is not the scoring criterion.

The six site profiles (Table~\ref{tab:sites}) map to real U.S. grid regions
with real coordinates. They are used to query the grounded data sources of
Appendix~\ref{app:data}. Two of the six, marked held-out, are never trained
on.

\begin{table}[H]
\centering\small
\caption{Site profiles and their real-world grid regions.}
\label{tab:sites}
\begin{tabular}{@{}llll@{}}
\toprule
Site key & State & Grid region (RTO/ISO) & Split \\
\midrule
tx\_erco    & Texas          & ERCOT & train \\
sd\_midwest & South Dakota   & MISO  & train \\
az\_west    & Arizona        & West (non-ISO) & train \\
va\_east    & Virginia       & PJM   & train \\
wa\_nw      & Washington     & West (non-ISO) & held-out \\
oh\_mid     & Ohio           & PJM   & held-out \\
\bottomrule
\end{tabular}
\end{table}

\section{Data Sources and Acknowledgments}
\label{app:data}
Every reported result is grounded in measured or officially projected data. We
do not use invented coefficients. Per-site water-stress level and its future
relative growth come from the WRI Aqueduct 4.0 baseline and future (SSP/RCP)
projections \citep{wri_aqueduct40}. Per-site grid carbon intensity and its
decarbonization slope come from EIA-930 hourly generation-by-fuel-type data.
Electricity price level, drift, and volatility come from EIA's industrial
retail price series \citep{eia930}. Source-dependent energization lead times
per site come from the LBNL interconnection-queue microdata
\citep{lbnl_queued_data}. On-site renewable capacity factor and its seasonal
volatility come from NREL PVWatts where reachable. Otherwise we fall back to a
solar-geometry estimate, and we report the bias explicitly
\citep{nrel_pvwatts}. Hourly temperature, a cooling-load driver, comes from
Open-Meteo \citep{openmeteo}.

The workload-efficiency (J/token) proxy is derived from device TDP and
throughput in the SemiAnalysis/InferenceX inference-benchmark dataset
\citep{semianalysis_inferencex}. It is cross-checked against an independent
market price and throughput signal from the Artificial Analysis API
\citep{artificialanalysis}. All six sites in Table~\ref{tab:sites} map to real
U.S. grid regions and states. A calibration pass (\texttt{data/calibrate.py}
in the released code) reports, per coefficient and per site, whether the value
was measured, projected, or fell back to a documented default. Every number in
this paper is therefore auditable. We thank WRI, LBNL, EIA, NREL, Open-Meteo,
SemiAnalysis, and Artificial Analysis for making this data available. None of
these organizations endorse this work.

\section{Limitations}
\label{app:limitations}
The market-scale figures motivating this work are third-party industry
projections
\citep{mckinsey2025,iea2025energyai,lbnl2025queued,westwater2025,
cbre2025capex,jll2026outlook}. They establish the magnitude and cadence of the
decision problem. They are not inputs to any reported result, and no empirical
claim in this paper depends on their precision. Reported per-MW build costs
vary by roughly an order of magnitude, depending on whether compute hardware
is included \citep{jll2026outlook}. We therefore use them to set the relative
scale of the switching-cost coefficient. We do not use them as an absolute
capital estimate.

The workload-energy signal is a throughput and device-power proxy. It is not
measured wall power. It informs relative efficiency trends and not absolute
energy accounting. Interconnection-queue data characterizes regional
generation headroom. It does not describe a specific facility's demand-side
connection. We use it as a supply-scarcity proxy, and the released
calibration code states this interpretation throughout.

The LCM is measurably outperformed by persistence on water stress
(Section~\ref{sec:results}, Figure~\ref{fig:wm}). It should not be used for
model-based planning on that axis until it clears that bar. The
\textsc{commit-once} front and learning-curve figures
(Figures~\ref{fig:fronts_co} and \ref{fig:learning}) each show a single
representative seed. Multi-seed hypervolume and sparsity aggregates for this
regime are now reported in Table~\ref{tab:main}. Held-out cost$\times$carbon
hypervolume is far below its in-distribution counterpart for every
configuration (Appendix~\ref{app:figures}). Frontiers at unseen sites should
therefore be treated as indicative and not decision-grade. Results use three
seeds, so reported standard deviations are themselves uncertain.
Static-scoring, myopic-greedy, and random baselines are specified in the
released code (\texttt{baselines/}). They remain in progress as of this draft.

Most fundamentally, policies whose consequences unfold over decades cannot be
validated against realized outcomes within a research cycle. We rely on
held-out trajectory evaluation and rolling-origin LCM backtesting. In ongoing
work, we are also running a coefficient sensitivity analysis. It samples each
grounded coefficient across its plausible range. We make no claim of verified
multi-decade forecasts.

\end{document}